\documentclass[10pt,letterpaper,conference]{IEEEtran}
\IEEEoverridecommandlockouts
\usepackage{cite}
\usepackage{amsmath,amssymb,amsfonts}
\usepackage{algorithmic}
\usepackage{graphicx}
\usepackage{textcomp}
\usepackage{xcolor}
\usepackage{geometry}
\usepackage[auth-lg,affil-it]{authblk}
\usepackage{graphicx}
\usepackage{booktabs}
\usepackage{textcomp}
\usepackage{xcolor}
\usepackage{tikz}
\usepackage{enumitem}
\usepackage{multirow}
\usepackage{amsmath}
\usepackage{graphicx}
\usepackage{subfigure}
\usepackage{makecell}
\usepackage{amsmath}
\usepackage{amssymb}
\usepackage{mathtools}
\usepackage{amsthm}
\usepackage{hyperref}
\usepackage[bottom]{footmisc}
\usepackage{tcolorbox}
\usepackage{hyperref}

\usepackage{dblfloatfix}    % To enable figures at the bottom of page
\usepackage[]{graphicx} % [demo] option for empty figure

\usepackage{xcolor}
\usepackage{fancyvrb}
\usepackage[]{footmisc}
\usepackage{xspace}
\definecolor{darkgreen}{rgb}{0.0, 0.5, 0.0}
\usepackage{cite}
\usepackage{amsmath,amssymb,amsfonts}
\usepackage{algorithmic}
\usepackage{graphicx}
\usepackage{textcomp}
\usepackage{xcolor}
\def\BibTeX{{\rm B\kern-.05em{\sc i\kern-.025em b}\kern-.08em
    T\kern-.1667em\lower.7ex\hbox{E}\kern-.125emX}}

\newcommand{\mixtral}{Mixtral\xspace}
\newcommand{\mamba}{BlackMamba\xspace}
\newcommand{\he}{HE\xspace}
\newcommand{\gs}{GS\xspace}
\newcommand{\cs}{CS\xspace}
\newcommand{\mt}{MATH\xspace}

\newcommand{\OBONE}{{\fontfamily{lmss}\selectfont Takeaway 1}}
\newcommand{\OBTWO}{{\fontfamily{lmss}\selectfont Takeaway 2}}
\newcommand{\OBTHREE}{{\fontfamily{lmss}\selectfont Takeaway 3}}
\newcommand{\OBFOUR}{{\fontfamily{lmss}\selectfont Takeaway 4}}
\newcommand{\OBFIVE}{{\fontfamily{lmss}\selectfont Takeaway 5}}
\newcommand{\OBSIX}{{\fontfamily{lmss}\selectfont Takeaway 6}}

\usepackage{tikz}
\usepackage{xcolor}

\begin{document}

\title{Understanding the Performance and Estimating the Cost of LLM Fine-Tuning}

%% DO NOT EDIT THE FOLLOWING

% \renewcommand\Authsep{\qquad}
% \renewcommand\Authand{\qquad}
% \renewcommand\Authands{\qquad}

\author[ ]{
    Yuchen Xia$^1$ \hspace{2em} 
    Jiho Kim$^2$ \hspace{2em} 
    Yuhan Chen$^1$ \hspace{2em}
    Haojie Ye$^1$ \hspace{2em}
    Souvik Kundu$^{3}$  \\ 
    Cong (Callie) Hao$^2$ \hspace{2em} 
    Nishil Talati$^1$
}

\affil[ ]{$^1$University of Michigan\hspace{2em}  $^2$Georgia Institute of Technology\hspace{2em}  $^3$Intel Labs}

\maketitle
%\thefootnote\relax\footnotetext{ \scriptsize$^{\S}$ Equal contribution.}
%% EDIT YOUR PAPER'S CONTENTS BELOW

% \thispagestyle{plain}
% \pagestyle{plain}
\pagestyle{empty}
\begin{abstract}
Due to the cost-prohibitive nature of training Large Language Models (LLMs), fine-tuning has emerged as an attractive alternative for specializing LLMs for specific tasks using limited compute resources in a cost-effective manner. 
In this paper, we characterize sparse Mixture of Experts (MoE) based LLM fine-tuning to understand their accuracy and runtime performance on a single GPU. 
Our evaluation provides unique insights into the training efficacy of sparse and dense versions of MoE models, as well as their runtime characteristics, including maximum batch size, execution time breakdown, end-to-end throughput, GPU hardware utilization, and load distribution. 
Our study identifies the optimization of the MoE layer as crucial for further improving the performance of LLM fine-tuning. 
Using our profiling results, we also develop and validate an analytical model to estimate the cost of LLM fine-tuning on the cloud. 
This model, based on parameters of the model and GPU architecture, estimates LLM throughput and the cost of training, aiding practitioners in industry and academia to budget the cost of fine-tuning a specific model.
\end{abstract}
\section{Introduction}
% \textcolor{red}{TODO: add references.}
Large Language Models (LLMs) are widely utilized in Natural Language Processing (NLP)~\cite{wei2022emergent}. 
Modern LLMs typically possess billions to trillions of parameters, necessitating extensive time and resources for training. 
For instance, the estimated cost of training OpenAI's GPT-4 model exceeds \$100 million, rendering it financially prohibitive for most small-to-medium size enterprises and the academic community~\cite{zhang2023dissecting}.
Given the open-sourcing of numerous pre-trained LLMs (e.g., LLAMA~\cite{touvron2023llama} and Mixtral~\cite{jiang2024mixtral}), fine-tuning has emerged as an attractive alternative for further specializing these models in a cost-effective manner~\cite{chung2022scaling}.
Given the learning ability of pre-trained models, it is feasible to use a domain-specific dataset to align the desired behaviors of LLMs through supervised fine-tuning on instruction-following tasks~\cite{lialin2023scaling}. 
Unlike pre-training, fine-tuning can be conducted in a resource-constrained environment, typically using one or a few GPUs. 
Consequently, fine-tuning presents a compelling case for applications such as specialized question answering within enterprises, legal document analysis and drafting, healthcare/medical research, technical and IT support, among others~\cite{chen2023parameterefficient}.

This paper characterizes LLM fine-tuning with two primary objectives: (1) understanding the performance characteristics of LLM fine-tuning, and (2) developing an analytical model to estimate the cost of fine-tuning on the cloud. 
Given our focus on cost-efficient LLM fine-tuning, we concentrate on fine-tuning sparse Mixture-of-Expert (MoE) models. 
Specifically, we employ an attention-based MoE model, \mixtral~\cite{jiang2024mixtral}, and a state-space MoE model, \mamba~\cite{anthony2024blackmamba}. 
Using these models and two domain-specific datasets for mathematics and common-sense question-answering, we conduct an in-depth profiling study to understand their performance characteristics with a single GPU. 
We compare the dense and sparse counterparts of the investigated MoE models to evaluate their learning rates and runtime performance. 
Our investigation covers memory consumption, maximum batch size supported within a single GPU memory budget, execution time breakdown and bottlenecks, overall throughput, microarchitectural performance counters, and runtime load distribution. 
The insights gained from our study are used to develop and validate an analytical model to estimate the cost.

Our characterization uncovers the following unique insights.
(1) Fine-tuning can be achieved in less than 10 epochs, and sparse MoE model that activates a subset of experts can learn as well as its dense counterparts.
(2) MoE layer consumes the highest fraction of execution time in LLM fine-tuning; optimizing MoE layer performance is key to improving the overall cost of LLM fine-tuning.
(3) Sparse MoE model improves end-to-end throughput by supporting a larger batch size.
Given similar learning abilities of sparse and dense models, it is desired to use a sparse MoE model for cost-effective fine-tuning.
(4) The workload becomes compute bound by increasing batch size; improving compute resources will increase performance.
(5) Fine-tuning sparse model leads to more load imbalance.

Based on these insights, we create an analytical model to estimate the cost of LLM fine-tuning based on model size, dataset size, and GPU architecture. First, we estimate the maximum batch size for a given GPU memory, then compute fine-tuning throughput. We validate this throughput with experimental results, showing an RMSE of less than 0.55. Using the estimated throughput, our model calculates the fine-tuning cost for different cloud providers.
% Using the above insights, we build an analytical model to estimate the cost of LLM fine-tuning given model and dataset sizes, and GPU architecture.
% Specifically, we first estimate the maximum batch size supported by a given GPU memory capacity, and then compute fine-tuning throughput using our model.
% We validate the throughput computed by our model with experimental results to show a low Root Mean Squared Error (RMSE) of less than 0.55.
% Using the estimated throughout, our model output the dollar-amount to fine-tune an LLM for different cloud providers.

The contributions of this paper are as follows.
\begin{itemize}
    \item Make a case for LLM fine-tuning for specializing pre-trained models in a cost-effective manner.
    \item A detailed accuracy and runtime performance analysis to understand the LLM fine-tuning workload behavior.
    \item Design and validation of an analytical model to estimate the cost of LLM fine-tuning in the cloud.
\end{itemize}
\section{Background}
\label{section:Preliminary}

\begin{figure}[t]
\centering
\centering
\includegraphics[width=\linewidth]{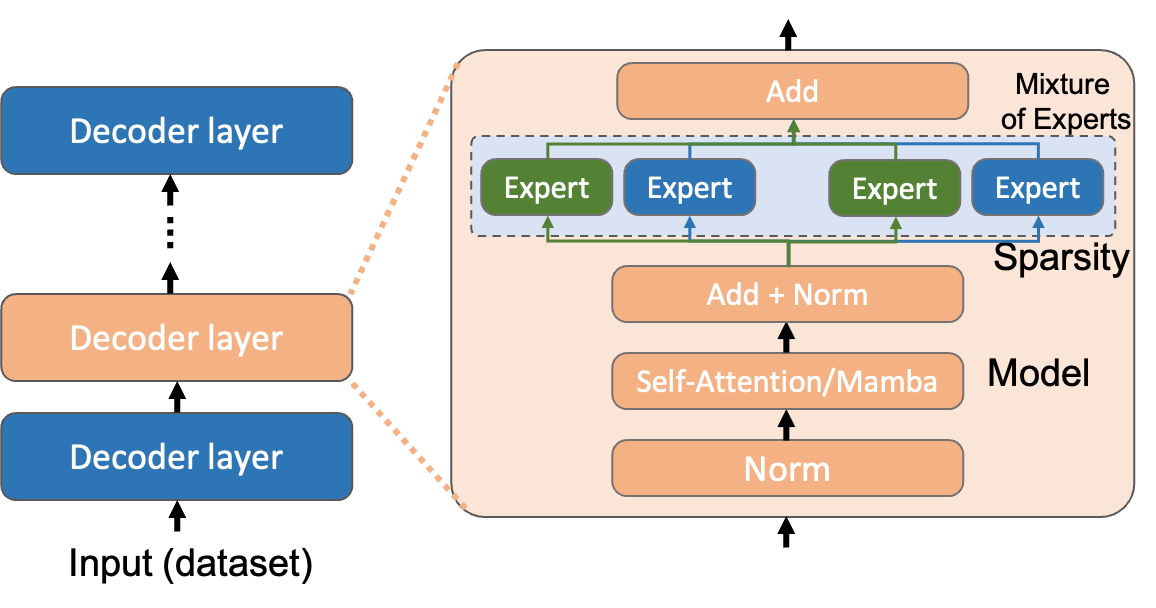}
\caption{LLM model overview. We evaluate accuracy, throughput, runtime, and GPU characterization for different models, input datasets, and fine-tuning sparsity. The different colored expert boxes in MoE layer means different sets of experts are activated according to the input token.}
\label{fig:overview}
\end{figure}

\subsection{LLM and Finetuning}
% \textcolor{red}{Explain the figure and LLM here...}
The decoder-only Transformer is designed to handle tasks where the output generation depends solely on the preceding tokens, making it particularly suited for auto-regressive tasks such as language modeling and text generation~\cite{radford2018improving}. 
In the classic decoder-only Transformer design, multiple decoder layers are connected in sequence. 
Each decoder layer consists of a self-attention block followed by a feed-forward network (FFN). 
Fig.~\ref{fig:overview} presents an overview of the decoder-only Transformer model with a Mixture-of-Experts (MoE) design. 
In this model, the FFN layers are divided into several smaller FFNs, referred to as experts, which are sparsely activated by a gating mechanism. 
The self-attention block can also be replaced with a Mamba layer to improve performance in sequence modeling (a model known as state-space model). 
LLMs like GPT~\cite{chatgpt, openai2024gpt4}, LLaMA~\cite{touvron2023llama}, Claude~\cite{claude}, Mistral~\cite{jiang2023mistral} have demonstrated their ability to excel in many natural language processing (NLP) tasks
% such as math and commonsense~\cite{davis2024mathematics}.
Training an LLM model from scratch requires a large amount of hardware resources and budget.

Fine-tuning LLMs allows organizations to harness the full potential of advanced AI systems by tailoring them to specific tasks and domains. 
This customization involves training the model on domain-specific data, enabling it to understand and generate content that aligns closely with the unique needs of the users. 
For instance, in the healthcare sector, a fine-tuned LLM can assist in diagnosing conditions by interpreting patient data and medical literature with high precision. 
Another attractive feature of fine-tuning LLMs is that it can be achieved at a cost-efficient manner.
While pre-training LLMs require thousands of GPU hours, fine-tuning can be achieved using a handful of GPUs in a relatively short amount of time~\cite{lialin2023scaling}.
This work uses case study of mathematics and common-sense question-answer datasets to demonstrate the fine-tuning process of LLMs.

% Fine-tuning is a technique that takes a pretrained LLM model and finetunes it on a small or task-specific dataset~\cite{wei2022}. The pretrained model is trained on large datasets that perform a wide range of tasks, while the finetuned LLM is for specialized domain-specific tasks and performs better than the pretrained general model on these tasks~\cite{houlsby2019parameterefficient}.  

\subsection{LoRA}
Low-Rank Adaption (LoRA) is a technique that freezes the pre-trained model weights and injects trainable rank decomposition into layers of the transformer architecture~\cite{hu2021}. 
LoRA significantly reduces the number of parameters, thereby decreasing the GPU memory footprint. 
LoRA can be used independently of the aforementioned fine-tuning techniques. 
In this work, we apply QLoRA~\cite{dettmers2023qlora} to the \mixtral-8x7B model~\cite{jiang2024mixtral}; more details are provided in \S\ref{sec:setup}.

% Low-Rank Adaption (LoRA) is a technique that freezes the pre-trained model weights, and injects trainable rank decomposition to layers of the transformer architecture~\cite{hu2021}. 
% LoRA greatly reduces the number of parameters, thus reducing the GPU memory footprint. 
% LoRA can be used independently with the aforementioned finetuning technique. 
% In this work, we apply QLoRA~\cite{dettmers2023qlora} to the \mixtral-8x7B model~\cite{jiang2024mixtral}, more details are in \S~\ref{sec:setup}.     

\subsection{Mixture of Experts (MoE)}
The quality of an LLM is highly related to its scale. 
Given a fixed computation budget, it is often desirable to train a model with more parameters to achieve higher accuracy.
Mixture-of-Experts (MoE) is a technique that, instead of using one large model for all tasks, combines multiple expert sub-networks into a single, large model.
As shown in Fig.~\ref{fig:overview}, with MoE, different sets of experts are selectively activated for different tokens.
This approach can significantly reduce the amount of computation required for both training and inference, enabling the scaling up of model size and achieving better model accuracy~\cite{asplos_keynote}.
\section{Experimental Setup} \label{sec:setup}
% Models: Mixtral8x7B, BlackMamba

% Datasets: Math\_14 for math training, GSM8k for math testing. commonsense\_15k for commonsense training, Hellaswag for commonsense testing. \BLUE{Ref to LLMAdaptor}

\begin{table}[]
    \centering
    \caption{LLM Models}
    \begin{tabular}{c|c|c|c|c}
         & \#params & Mem consump. & \#layers & \#MoE layer \\
         \hline \hline
         \mixtral & 47B & 23.35GB & 32 & 8 \\
         \mamba & 2.8B & 5.6GB & 18 & 8 \\
    \end{tabular}
    \vspace{5pt}
    \label{tab:llm_model}
\end{table}

\begin{table}[]
    \centering
    \caption{Datasets}
    \begin{tabular}{c|c|c|c}
         & \#queries & m. seq len & type \\
         \hline \hline
         Commonsense\_15K (\cs) & 15K & 79 & Common Sense \\ 
         Math\_14K (\mt) & 14K & 174 & Math \\ 
         Hellaswag (\he) & 10K & 272 & Common Sense \\ 
         GSM8K (\gs) & 1.3K & 148 & Math \\ 
    \end{tabular}
    \vspace{5pt}
    \label{tab:datasets}
\end{table}

\textbf{Models.}
We fine-tune two pre-trained MoE models, Mixtral-8x7B (\mixtral for short)~\cite{jiang2024mixtral} and BlackMamba-630M/2.8B (\mamba for short)~\cite{anthony2024blackmamba}.
The details of these models are shown in Table~\ref{tab:llm_model}.
Both models incorporate eight experts in their MoE layers.
For dense fine-tuning, all experts are activated, whereas for sparse fine-tuning, only the top two experts are selected for each token.

These models differ significantly in their transformer architectures and sizes. 
\mixtral is a conventional MoE transformer model with a total of 47 billion parameters. 
In contrast, \mamba is a state-space model that replaces all attention layers with mamba layers and has only 2.8 billion parameters. 
We fine-tune the full \mamba model (i.e., original weight matrices), whereas employed QLoRA~\cite{dettmers2023qlora} for parameter-efficient fine-tuning (PEFT) on \mixtral due to GPU memory capacity budget.
For QLoRA, we target the MoE layers, including the routers, and set the rank of the LoRA modules to 16. 
We enable FlashAttention2~\cite{dao2023flashattention2} during \mixtral fine-tuning for enhanced efficiency. 
Moreover, we use gradient checkpointing~\cite{chen2016training} to save memory usage.

\textbf{Datasets.}
Our fine-tuning process is implemented in PyTorch using the LLaMA-Factory framework~\cite{zheng2024llamafactory}, with a learning rate of 5e-5 and 10 epochs. 
Both models were fine-tuned on two datasets focused on different tasks: commonsense\_15k (\cs) and Math\_14k (\mt), which address commonsense reasoning and arithmetic reasoning respectively (provided by LLM-adapters~\cite{hu2023llmadapters}).
The details of datasets are used in Table~\ref{tab:datasets}.
For evaluation, we tested the models on GSM8K~\cite{cobbe2021gsm8k} for arithmetic reasoning and \he~\cite{zellers2019hellaswag} for commonsense reasoning.
Each dataset consists of thousands of queries. 
We define a query as the concatenation of a prompt and its ground-truth answer, which is feed to LLMs for fine-tuning.

\textbf{Profiling experiments.}
We evaluate the fine-tuning process from both software and hardware perspectives.
The software evaluation includes an end-to-end assessment of the fine-tuning process and measures the performance of the two models on various tasks post-fine-tuning.
Using PyTorch, we provide essential algorithm-level information such as test accuracy, training throughput, and layer-level latency breakdown.
The hardware evaluation offers a detailed analysis of GPU performance. 
Utilizing NVIDIA Nsight Compute~\cite{nsightcompute}, we gather kernel-level information, including SM utilization, memory utilization, and kernel latency. 
These metrics collectively offer a comprehensive overview of the models' performance, capturing both high-level algorithmic efficiency and detailed hardware utilization. 
Software evaluation is dataset-dependent, and we will show the test accuracy and fine-tuning throughput by utilizing both datasets. 
In contrast, hardware evaluation is dataset-independent as these workload characteristics do not depend on runtime data.
Because profiling is time-consuming (approximately 10,000$\times$ costlier compared to a native run without the profiler enabled), we manually set the batch size and sequence length to facilitate a more direct and efficient profiling process.

\begin{figure}[t]
\centering
\includegraphics[width=\linewidth]{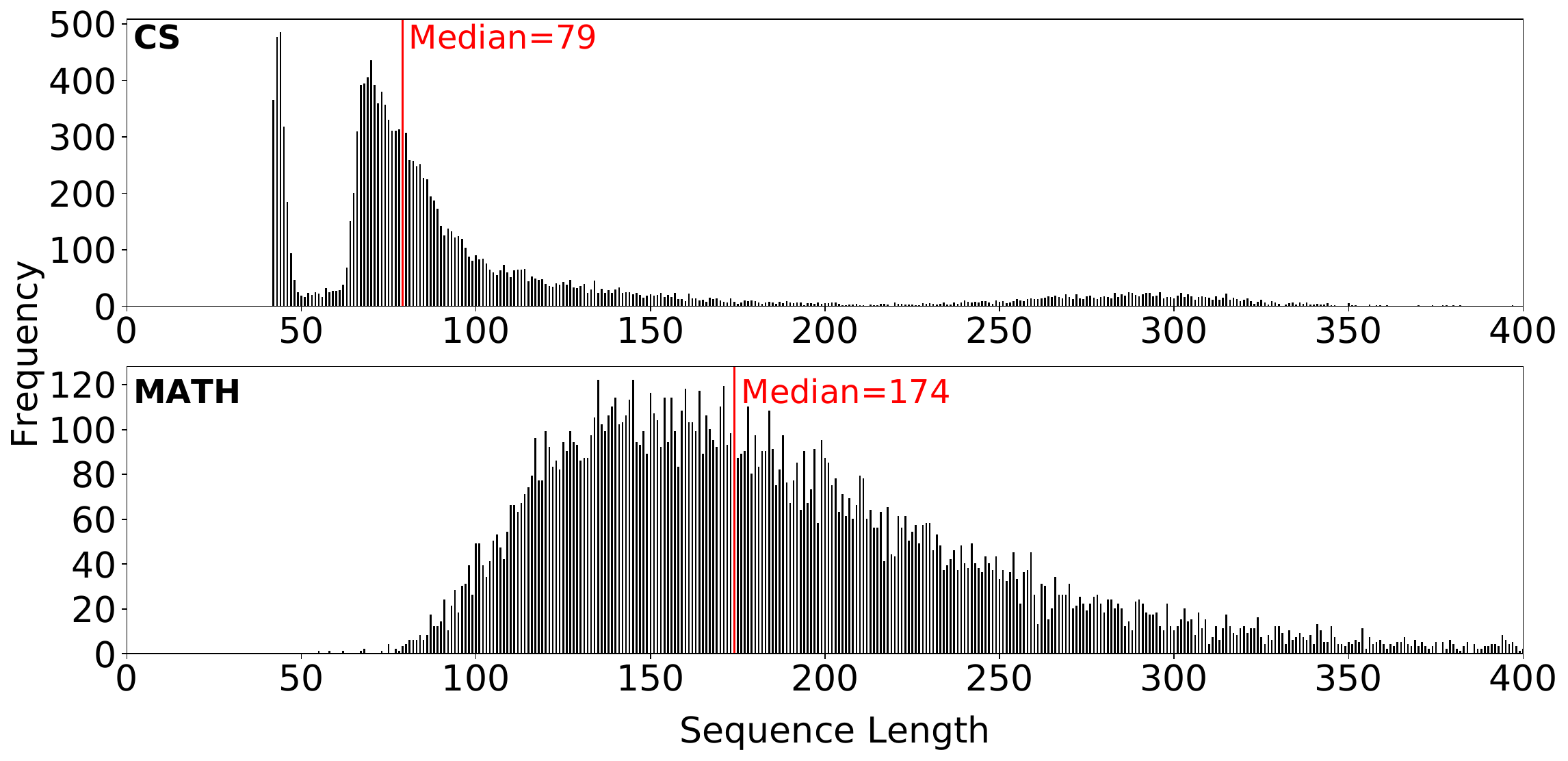}
\caption{Sequence length distribution for evaluated datasets.}
\label{fig:token_length}
\end{figure}
We present the sequence length distribution for the \cs and \mt datasets in Fig.~\ref{fig:token_length}. 
The median sequence length is 79 for \cs and 174 for \mt. 
Therefore, we select a sequence length of 128 for the hardware evaluation section to achieve an approximate profiling effect.
We also show a sensitivity study by varying sequence length to demonstrate its effect on performance.

\textbf{GPU platform.}
Our study is focused on characterizing the LLM fine-tuning process on a resource-constrained environment.
Therefore, we focus on fine-tuning these models on a single GPU.
Specifically, we conduct our experiments using NVIDIA A40 GPU with Ampere architecture.
The GPU has 48GB memory.
\textit{While our profiling study is based on this particular GPU, we show the versatility of our analytical model by validating our model against three other GPU with different sizes of compute and memory resources: (1) A100 GPU with 40GB memory, (2) A100 GPU with 80GB memory, and (3) H100 GPU with 80GB memory.
}We use Python v3.8.10, PyTorch v2.1.0, and CUDA v11.8.

% We conduct all experiments
% All experiments were conducted on an NVIDIA A40 (Ampere) GPU with 48GB GDDR6 memory, utilizing Python v3.8.10, PyTorch v2.1.0, and CUDA v11.8.

\section{Characterization Study}
Using the experimental setup discussed above, next, we conduct an in-depth characterization of LLM fine-tuning to understand both accuracy and runtime behaviors.

\subsection{Analysis of Model Trainability}
\label{sec:accuracy}

\begin{figure}[t]
\centering
\includegraphics[width=0.9\linewidth]{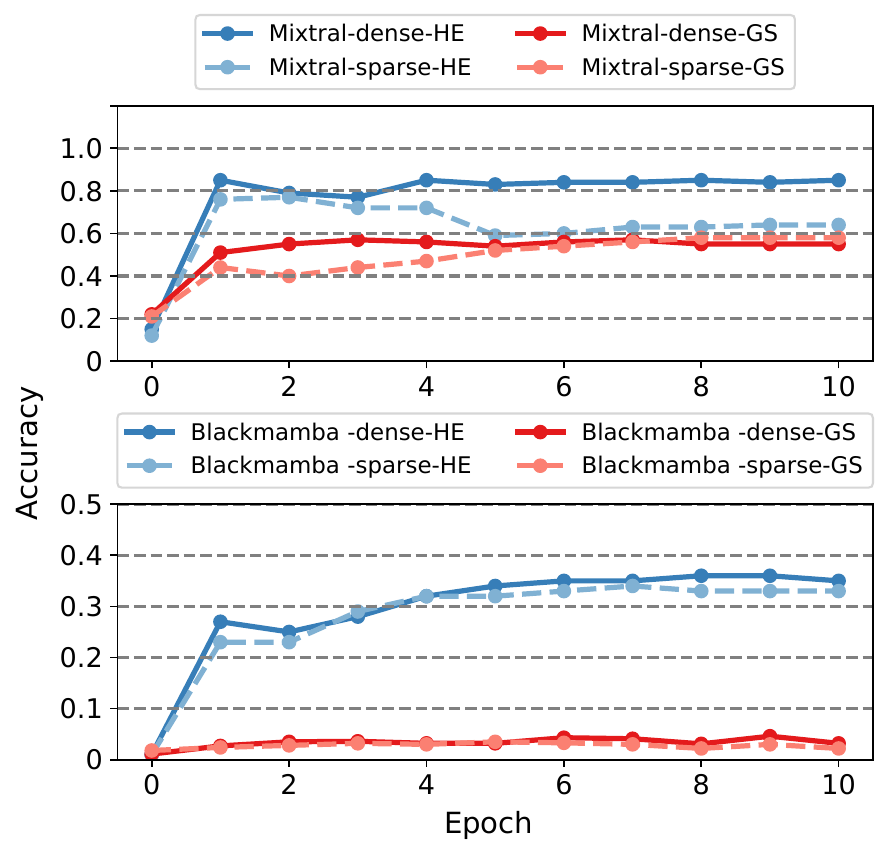}
\caption{Testing accuracy of \mixtral and \mamba. Both models are evaluated on two datasets Hellaswag (\he) and GSM8K (\gs), using dense and sparse fine-tuning.}
\label{fig:accuracy}
\end{figure}

We first evaluate if fine-tuning sparse LLM models can achieve the desired accuracy levels. Pre-trained models show low accuracy: \he and \gs have under 25\% on \mixtral and under 10\% on \mamba. We assess accuracy improvements post-fine-tuning and compare the learning capabilities of dense and sparse versions of both models.
% First, we aim to determine whether it is feasible to fine-tune these sparse LLM models to achieve desired accuracy levels. 
% The pre-trained models exhibit low accuracy on the datasets assessed in this study.
% Specifically, \he and \gs achieve less than 25\% accuracy on \mixtral and less than 10\% accuracy on \mamba. 
% Utilizing our setup, we assess the improvement in testing accuracy following fine-tuning of the models.
% Additionally, we compare the learning capabilities of the dense and sparse versions of both models upon fine-tuning.

Fig.~\ref{fig:accuracy} shows the testing accuracy of \mixtral and \mamba on two datasets Hellaswag (\he) and GSM8K (\gs).
We fine-tune both models using the sparse and dense setups described in \S\ref{sec:setup} for 10 epochs, and test the accuracy of the fine-tuned model at each epoch. 
We make the following observations in Fig.~\ref{fig:accuracy}. 
(1) Fine-tuning converges relatively quickly. 
Typically, 10 epochs are enough for fine-tune models to stabilize at or close to their peak accuracy.
On \gs, both models are close to their peak accuracy at the first epoch.
% \BLUE{(possible consensus from other work?)}
(2) The smaller model \mamba takes relatively more epochs to reach its peak accuracy, as it took \mamba 5 epochs to converge on HE. 
(3) The larger model \mixtral has better accuracy compared to \mamba on both datasets.
(4) Both models perform better on the \cs dataset HE than on the \gs dataset GS. 
This is because math is harder for smaller LLMs to learn~\cite{ahn2024large}.
The \mamba model is inadequate for fine-tuning \gs.
% The poor accuracy of \mamba on \gs means fine-tuning does not work well for \mamba to have a reasonable performance on math tasks. 
This is likely attributed to the complexity of mathematical tasks and the smaller model size of \mamba. Additionally, Mamba is specifically engineered for long sequence modeling, potentially resulting in unsatisfactory arithmetic reasoning ability~\cite{gu2024mamba}. 
Thus, in our characterization study in later sections, we will not show the results for \mamba fine-tuned on \mt. 
(5) The performance of sparse fine-tuning is close to that of dense fine-tuning, with the exception of \mixtral on \he. However, even for this outlier, sparse fine-tuning achieves similar peak accuracy compared to dense; we see a drop of accuracy between the epoch 4 and 5, and indicates sparse fine-tuning is more vulnerable to over-fitting, especially for easy tasks~\cite{xue2022student}. 
Following the above insights, the key take-away of this analysis can be summarized as follows.

\begin{tcolorbox}[width=0.48\textwidth]
\textbf{\OBONE.}
Sparse model can be trained as well as its dense counterpart.
% The accuracy of sparse fine-tuning can be as accurate as dense.

\textbf{\OBTWO.}
Fine-tuning generally takes less ten epochs to reach peak accuracy. 
\end{tcolorbox}

% We show that Mamba is not suitable for math (explain reason here), as shown in figure 
% \GREEN{(a figure for Mixtral and BlackMamba accuracy, showing blackmamba is very bad on GSM8K)}. Thus we'll only use Mistral for the following characterization on math. \RED{It seems Blackmamba is also bad on commonsense compared to Mixtral, is it simply because of the smaller model size?}

% \BLUE{Show the distribution of sequence length in training/testing datasets, and explain why we use the sequence length in the later ablation study. Better to use median or mean sequence length for non-ablation study}

\subsection{Analysis of Runtime Performance}
% After confirming that both \mixtral and \mamba can be fine-tuned to achieve a reasonable level of accuracy in multiple cases, we proceed to examine the performance characteristics of fine-tuning in a resource-constrained environment, specifically using only one GPU.
% This setup allows us to demonstrate unique runtime characteristics, including execution time breakdown, fine-tuning throughput, maximum batch size supported, compute and memory utilization, load imbalance, and sensitivity analysis.
% We also compare the performance of both sparse and dense models.
% The insights gained from this study will be utilized to develop and correlate a robust analytical model to estimate the cost of fine-tuning.
After confirming that both \mixtral and \mamba can be fine-tuned to achieve acceptable accuracy, we examine their performance in a resource-constrained environment using a single GPU. This setup highlights unique runtime characteristics such as execution time breakdown, throughput, maximum batch size, compute and memory utilization, load imbalance, and sensitivity analysis. We also compare sparse and dense models. Insights from this study will help develop a robust analytical model for estimating fine-tuning costs.
% \textcolor{red}{Write about what we do with the dataset!}

\subsubsection{Maximum Batch Size Support}
% \textcolor{red}{Yuchen}
The maximum batch size in fine-tuning is determined by GPU memory size, model size, sequence length, and MoE sparsity. 
The LLM occupies a certain amount of GPU memory, with the remainder available for intermediate data during fine-tuning. 
Longer sequence lengths consume more memory, and denser MoE configurations require additional memory space. 
We discuss the heuristic for determining the maximum batch size in \S\ref{sec:analytical}.
Based on our experimental study on NVIDIA A40 GPU with 48GB memory, we empirically find and report the maximum batch size supported by different model and dataset combinations in Table~\ref{tab:max_bs}.

% The maximum batch size in fine-tuning is determined by GPU memory size, model size, sequence length, and the MoE sparsity. 
% The LLM model tasks up a certain amount of memory in GPU, and the rest can be used for intermediate data in fine-tuning. The longer sequence length will take up more memory, and the denser the MoE is, the more memory space is needed. We discuss the heuristic to get the maximum batch size in \S\ref{sec:analytical}. Table~\ref{tab:max_bs} shows the maximum batch size in this work.

\begin{table}[]
    \centering
    \caption{Maximum batch size supported by LLM fine-tuning; D: dense and S:sparse.}
    \begin{tabular}{c|c|c|c|c}
          & \mixtral-D & \mixtral-S & \mamba-D & \mamba-S  \\
    \hline \hline
         \cs & 2 & 8 & 6 & 20 \\
         \mt & 1 & 3 & 2 & 8 \\ 
    \end{tabular}
    \vspace{5pt}
    
    \label{tab:max_bs}
\end{table}

\subsubsection{Execution Time Breakdown}
\begin{figure}[t]
\centering
\includegraphics[width=\linewidth]{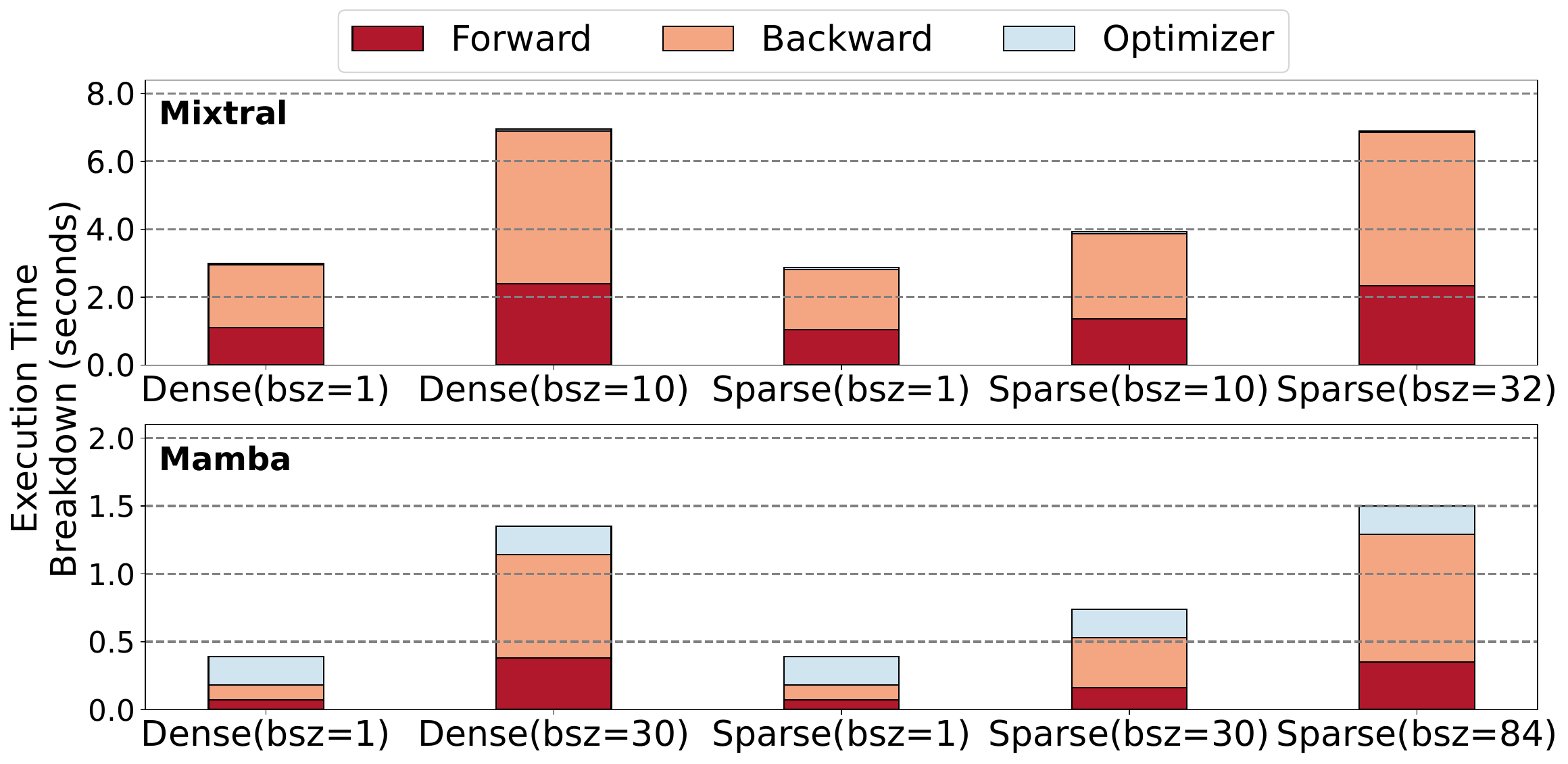}
\caption{Execution time breakdown.}
\label{fig:latency_highlevel}
\end{figure}
We first analyze the high-level execution time breakdown for \mixtral and \mamba.
The purpose of this study is to understand where does this workload spend most of its time.
% \textcolor{red}{why seq 128?}.
As discussed in \S\ref{sec:setup}, we conduct this study using a sequence length of 128.
% \GREEN{The median sequence length is 79 and 174 for \cs and \mt respectively. To eliminate the influence of sequence length in this study, we use an input sequence length of 128 for all model-dataset combinations, which falls between the two datasets.}

At a high-level, the fine-tuning workload can be divided into three stages: (1) forward, (2) backward, and (3) optimizer.
We use a batch size of 1 and the maximum batch size supported by a model-dataset combination to show workload characteristics.
Fig.~\ref{fig:latency_highlevel} illustrates the following insights.
(1) The optimizer stage in \mamba fine-tuning takes a considerable portion of the running time (up to 53\% when conducting sparse fine-tuning with batch size = 1), while the execution time share of the optimizer stage in \mixtral fine-tuning is negligible.
The running time of the optimizer stage depends only on the number of parameters that need to be updated during fine-tuning. 
This difference is primarily due to the different fine-tuning strategies applied to these two models: only the parameters in the LoRA module are updated for \mixtral fine-tuning, whereas \mamba undergoes full fine-tuning.
(2) The runtime of the forward and backward stages increases with sparsity and batch size due to the increased amount of computation.
(3) The backward stage typically takes more time than the forward stage. 
In \mamba, the backward stage demands more computation than the forward stage due to the need for gradient calculation and propagation, resulting in two matrix multiplication operations. In \mixtral fine-tuning, gradient calculation adds minimal computation as only a small portion of parameters need it. However, gradient checkpointing in \mixtral saves memory but increases the backward stage runtime due to the re-computation of intermediate values.
% For \mamba, the backward stage requires more computation than the forward stage. 
% For instance, the matrix multiplication operations need to calculate the gradient for the weight matrix and propagate the gradient to the next layer, resulting in two matrix multiplication operations in the backward stage. 
% In \mixtral fine-tuning, the gradient calculation does not add significantly to the computation in the backward stage because only a small portion of parameters associated with LoRA need gradient calculation.
% However, since gradient checkpointing is used in \mixtral fine-tuning to save memory, the re-computation of intermediate values results in a longer running time for the backward stage.

\begin{figure}[t]
\centering
\includegraphics[width=\linewidth]{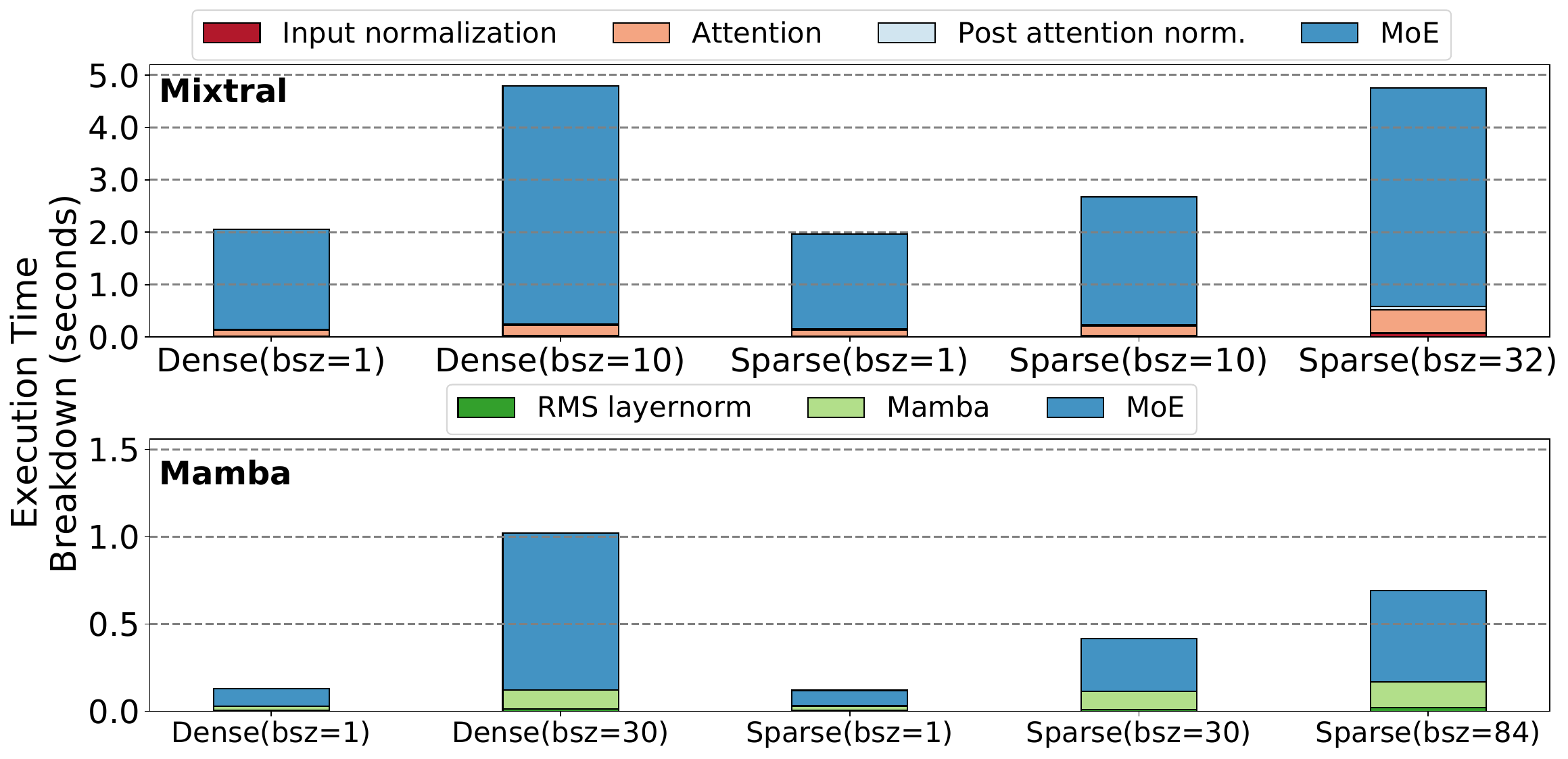}
\caption{Execution time breakdown in terms of different model layers. }
\label{fig:latency_layerlevel}
\end{figure}
We further investigate the execution breakdown based on various layers in two LLM models. 
For \mixtral, these layers include input normalization, attention, post-attention normalization, and MoE. 
In contrast, \mamba comprises the Mamba layer, Root Mean Squared (RMS) layer normalization, and MoE. As shown in Fig.~\ref{fig:latency_layerlevel}, the MoE layer is the most time-consuming, accounting for 85\% of the overall execution time on average. 
The execution time for the MoE layer encompasses both the forward and backward passes during fine-tuning. 
\textit{Consequently, MoE is the costliest layer and a prime target for optimization to enhance the performance of LLM fine-tuning.}

% As shown in Fig.~\ref{fig:latency_layerlevel}, the MoE layer accounts for at least 70\% and 90\% of the running time of the forward and backward stages in \mamba and \mixtral fine-tuning, respectively.

\begin{figure}[t]
\centering
\includegraphics[width=\linewidth]{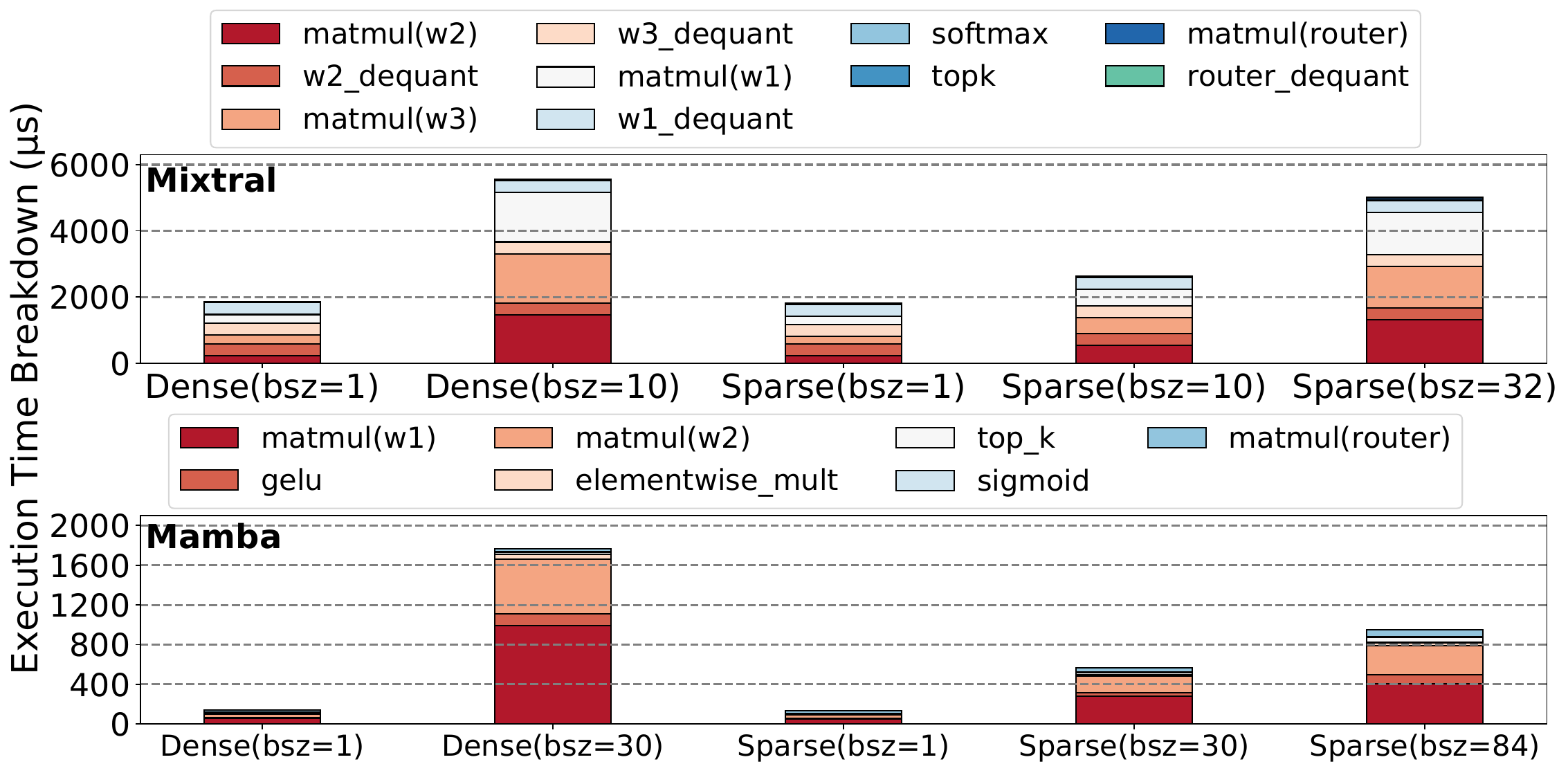}
\caption{Execution breakdown of the MoE layer for different kernels.}
\label{fig:latency_kernellevel}
\end{figure}
% \textcolor{red}{Add a small figure showing MoE architecture for both mamba and mixtral.}

\begin{figure}[t]
\centering
\includegraphics[width=\linewidth]{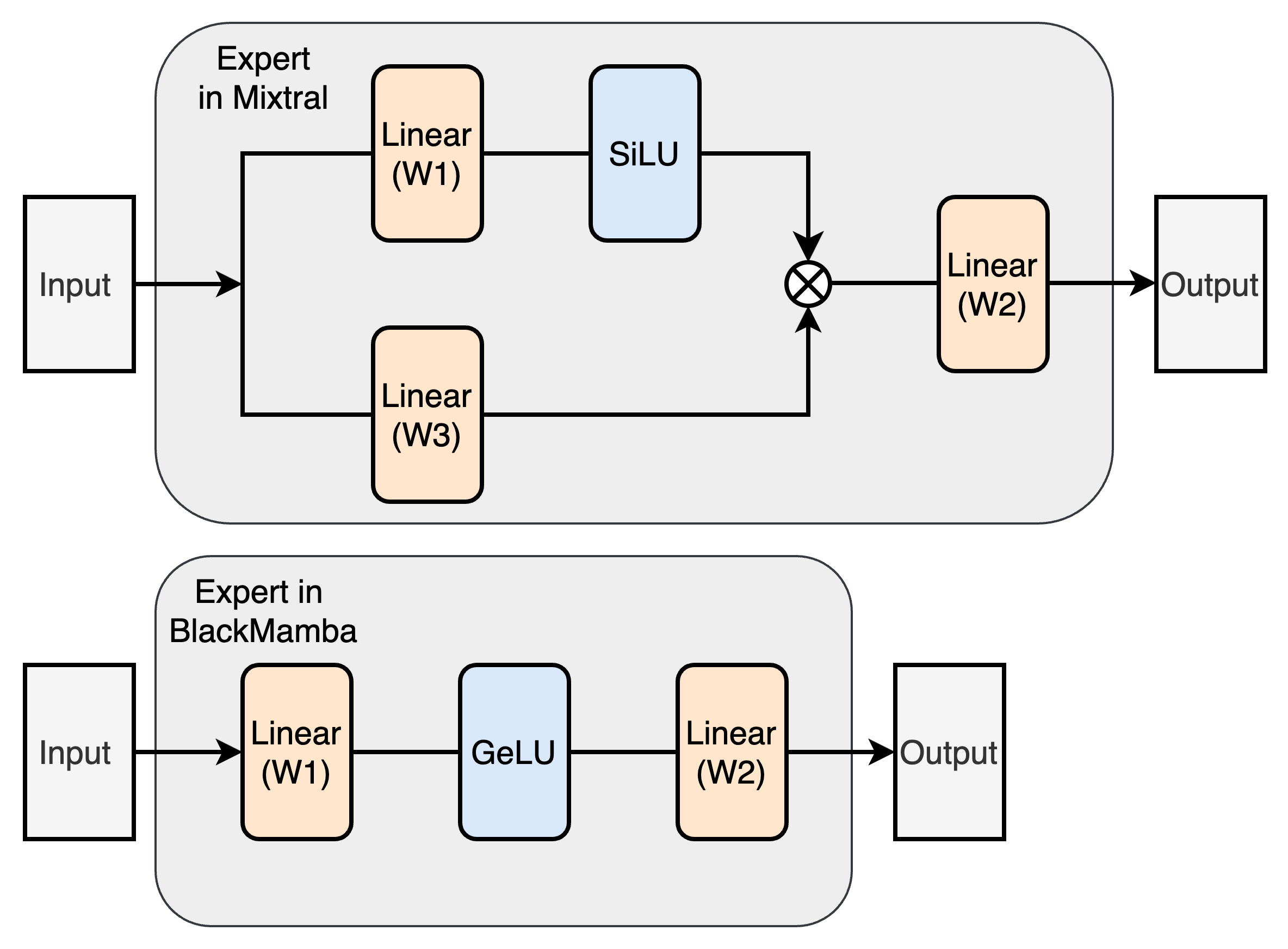}
\caption{Expert architectures for Mixtral (top) and BlackMamba (bottom).}
\label{fig:ffn_arch}
\end{figure}

To concretely understand the opportunity for improving MoE layer performance, we also perform a kernel-level analysis within the MoE layer. 
Fig.~\ref{fig:ffn_arch} illustrates the architecture of the MoE layer in both \mixtral and \mamba models. 
Each expert in \mamba consists of a standard Feed-Forward Network (FFN) layer with two serially connected weight matrices (W1 and W2) and a Gelu activation layer between. 
In contrast, experts in \mixtral are FFN layers with Swish-Gated Linear Units, involving an additional weight matrix (W3) in parallel with W1.

% To concretely understand the opportunity to improve MoE layer performance, we also perform a kernel-level analysis within the MoE layer.
% Fig.~\ref{fig:ffn_arch} shows the architecture of the MoE layer in both \mixtral and \mamba models.
% Each expert in the MoE layer of \mamba is a standard Feed-Forward Network (FFN) layer with two serially connected weight matrices (W1 and W2) and a Gelu activation layer between them. 
% In contrast, experts in \mixtral are FFN layers with Swish-Gated Linear Units that involve an additional weight matrix (W3) in parallel with W1.

Fig.~\ref{fig:latency_kernellevel} shows the kernel-level MoE time breakdown.
The figure clearly shows that matrix multiplication (W1, W2, and W3) is the largest component of the MoE layer for both \mamba and \mixtral. 
As batch size and sparsity increase, so does computational demand, prolonging matrix multiplication latency. The de-quantization operation in \mixtral fine-tuning also becomes significant, especially with low sparsity and small batch sizes. While quantization reduces model size and memory footprint, it can increase computation time due to de-quantization. This highlights the need to evaluate trade-offs between memory savings and computation time, particularly in scenarios with small batch sizes and sequence lengths.
% The total number of tokens in each batch rises with both sparsity and batch size, consequently escalating computational demand and prolonging matrix multiplication latency.
% Additionally, the runtime of the de-quantization operation for each weight matrix is significant in \mixtral fine-tuning, becoming dominant with low sparsity and small batch sizes. Quantization reduces model size and memory footprint but may introduce additional computational overhead due to dequantization during fine-tuning. 
% This suggests evaluating the performance trade-offs between reduced memory consumption and increased computation time, especially in scenarios characterized by small batch sizes and sequence lengths.
% \textcolor{red}{why?}
% \textcolor{red}{What is the message here? What do we need to optimize to best optimize fine-tuning performance?}

\begin{tcolorbox}[width=0.48\textwidth]
\textbf{\OBTHREE.}
Matrix multiplication operations in the MoE layer contribute significantly to the end-to-end execution time, making the MoE layer the costliest component in LLM fine-tuning.
% MoE layer is the most time-consuming, and in MoE matrix multiplication takes the majority of time.
\end{tcolorbox}

\subsubsection{Fine-Tuning Throughput}
\begin{figure*}[t]
\centering
\includegraphics[width=\linewidth]{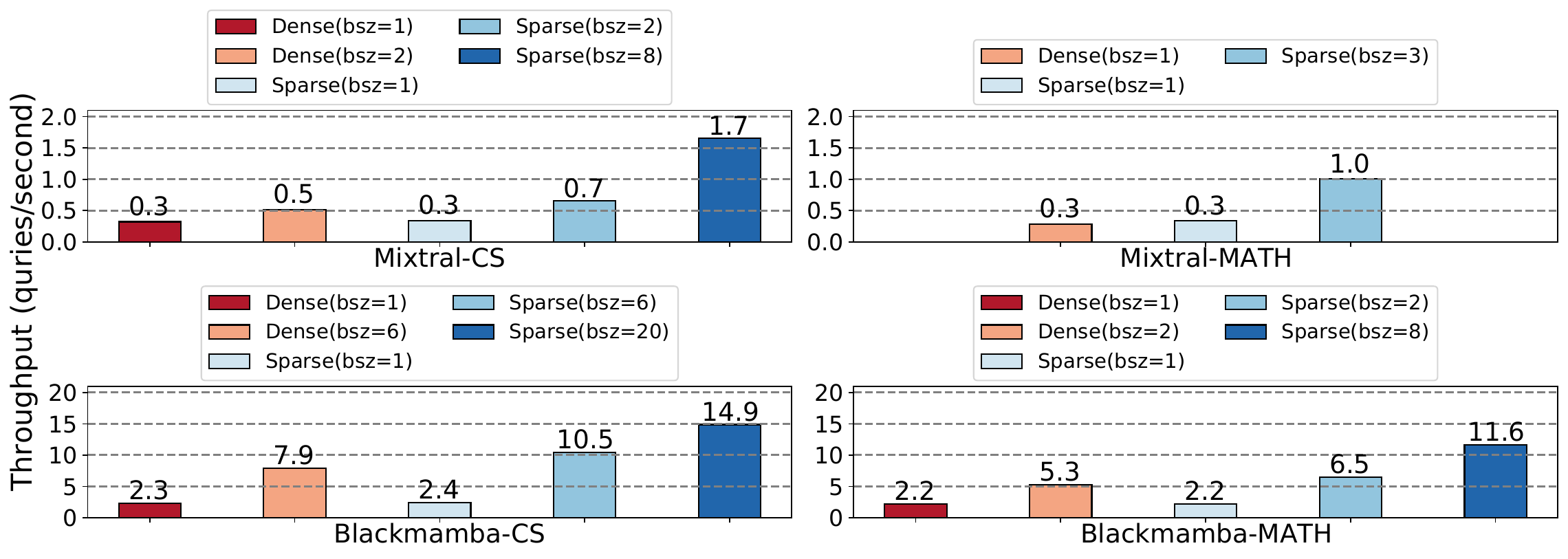}
% \vspace{-0.8cm}
\caption{Query throughput of \mixtral and \mamba.}
\label{fig:throughput}
\end{figure*}

\begin{figure*}[h!]
\centering
\includegraphics[width=\linewidth]{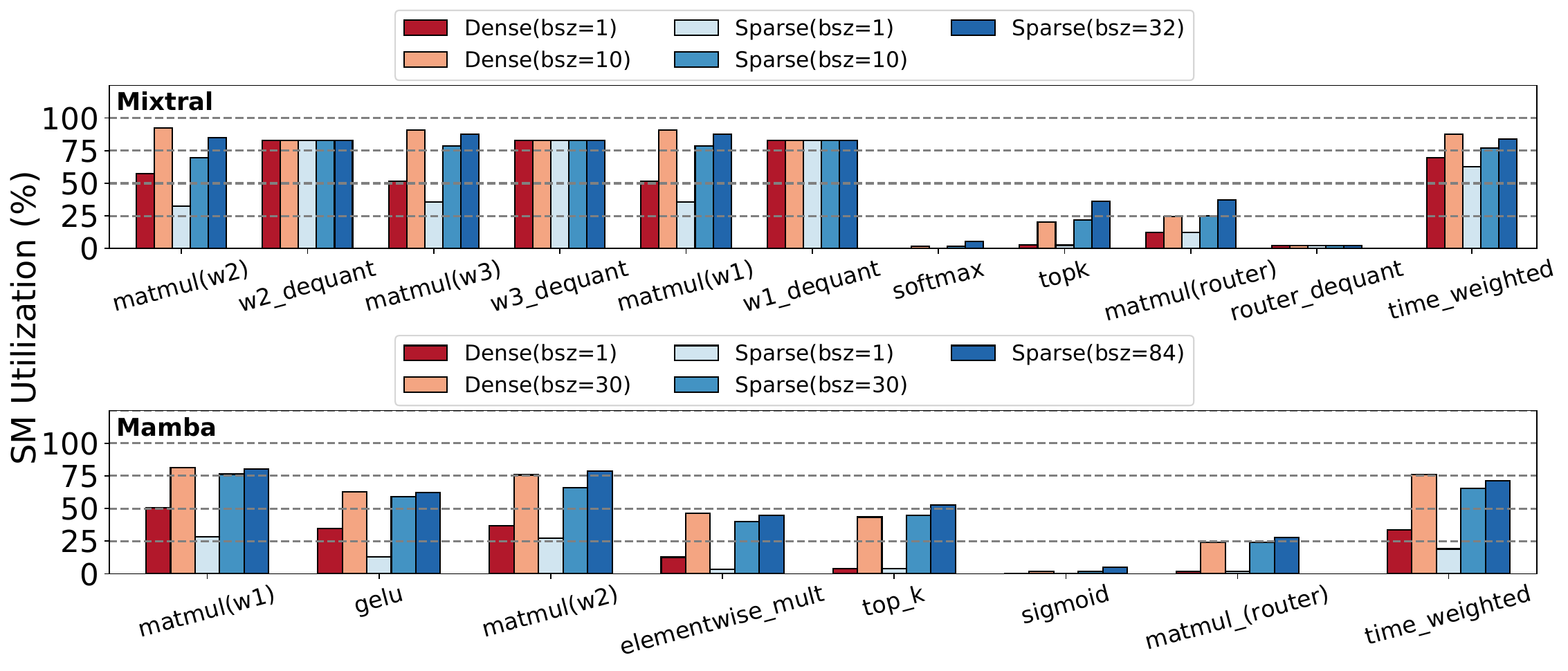}
% \vspace{-0.8cm}
\caption{GPU SM utilization of different kernels in the MoE layer for different batch sizes.}
\label{fig:sm_util}
\end{figure*}

\begin{figure*}[h!]
\centering
\includegraphics[width=\linewidth]{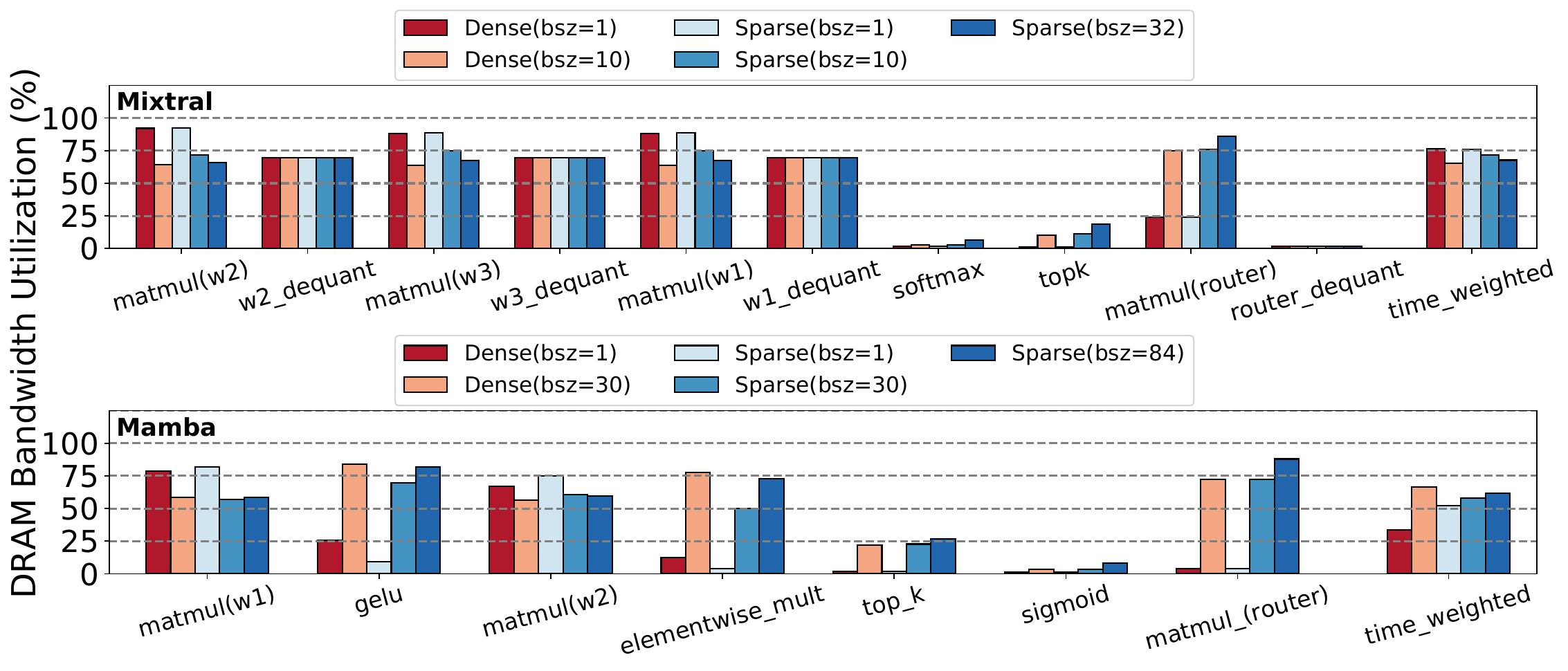}
% \vspace{-0.8cm}
\caption{GPU DRAM bandwidth utilization of different kernels in the MoE layer for different batch sizes.}
\label{fig:mem_util}
\end{figure*}
Next, we present the fine-tuning throughput of \mixtral and \mamba on the \mt and \cs datasets separately in Fig.~\ref{fig:throughput}.
We use a throughput metric of queries/second processed, where a query includes a prompt and a ground-truth answer for fine-tuning.
To obtain these results, we extract 1000 examples from each dataset and fine-tuned \mixtral and \mamba on them using the smallest batch size (batch size = 1) and the largest batch size that would fill the GPU memory.

As illustrated in Fig.~\ref{fig:throughput}, sparse fine-tuning achieves higher throughput than dense fine-tuning.
This is because the sparse fine-tuning baseline consumes less memory to store intermediate values, which allows for higher batch sizes compared to its dense counterpart.
% For example, the \mixtral model on \cs dataset can only support a batch size of 2 with a dense model, and a batch size of 8 with a sparse model that activates 2 out of 8 experts.
Additionally, with the same batch size, sparse fine-tuning achieves higher throughput because it involves fewer computational demands, resulting in lower latency.
This is evident when comparing the throughput of batch size of 2 in \mixtral-\cs for dense (0.5 qps) vs. sparse (0.7 qps) models.

Fig.~\ref{fig:throughput} also shows that throughput does not increase linearly with batch size. 
For instance, sparse fine-tuning of \mixtral-\cs improves throughput by 1.9$\times$ when increasing the batch size from 1 to 2, but only by 4.8$\times$ when increasing from 1 to 8. 
With smaller batch sizes, the SM utilization rate is lower, providing enough computational resources to feed more operations in parallel. 
However, as the batch size continues to increase, the SMs become saturated (more details in \S\ref{sec:GPUcharacterization}), and we can no longer hide latency by better utilizing computational resources.

% Additionally, Fig.~\ref{fig:throughput} demonstrates that \mamba achieves a much higher throughput on the same dataset compared to \mixtral (although, the accuracy levels are different as shown in Fig.~\ref{fig:accuracy}). 
% \mamba's smaller size allows for a larger batch size, and it has lower latency compared to \mixtral at the same batch size. 
% As shown in \S\ref{sec:setup}, the average sequence length in the \cs dataset is smaller than in the \gs dataset, allowing both \mixtral and \mamba to have larger batch sizes during fine-tuning on \cs, resulting in higher throughput.

\begin{tcolorbox}[width=0.48\textwidth]
\textbf{\OBFOUR.}
Sparse model significantly improves throughput, reducing end-to-end cost of fine-tuning.
\end{tcolorbox}

\subsubsection{Hardware characterization}
\label{sec:GPUcharacterization}

As shown in Fig.~\ref{fig:latency_highlevel}, the execution time of LLM fine-tuning is dominated by the MoE layer.
To offer further insights, we use detailed microarchitecture  hardware metrics on the GPU to further understand execution bottlenecks in the MoE layer.
The goal of this study is to identify whether various kernels in the MoE layers are bound by compute or memory resources, and how future GPU designs can further scale performance.

\textbf{Compute resource utilization study.}
Fig.~\ref{fig:sm_util} shows the kernel-level breakdown of GPU Streaming Multi-processor (SM) utilization for the MoE layer.
This utilization is weighted by the amount of time each kernel takes.
We use a sequence length of 128 (\S\ref{sec:setup}).
Sequence length will influence the choice of batch size, and we discuss the effects of sequence length on runtime, throughput, SM utilization, and memory utilization in \S\ref{sec:sensitivity}. 
For dense fine-tuning, we show the SM utilization of batch size 1 and the maximum batch size that fits into memory; for sparse fine-tuning, we use the two batch sizes for dense fine-tuning, and the maximum batch size that fits into memory.

Fig.~\ref{fig:sm_util} shows the SM utilization of different kernels in the MoE layer, which offers the following insights.
(1) For both sparse and dense fine-tuning, SM utilization increases with batch size due to higher parallelism and GPU activity.
(2) Sparse fine-tuning has lower SM utilization than dense fine-tuning at the same batch size because it activates only 2 out of 8 experts, reducing parallelism. Consequently, sparse fine-tuning supports a higher maximum batch size. Both achieve similar maximum SM utilization at their peak batch sizes.
(3) The de-quantization kernel maintains high SM utilization regardless of batch size.
(4) Matrix multiplication kernels achieve higher SM utilization with larger batch sizes, leveraging the GPU's parallel processing capabilities.
% (1) For both sparse and dense fine-tuning, the SM utilization increases with increasing batch size. 
% This is because a higher batch size can achieve higher parallelism, and keep GPU busy. 
% (2) With the same batch size, sparse fine-tuning has a lower SM utilization compared to dense fine-tuning.
% This is because sparse fine-tuning activates only 2 out of the 8 experts, resulting in lower parallelism. 
% Due to the same reason, sparse fine-tuning has a higher maximum batch size than that of dense fine-tuning. 
% Both sparse and dense fine-tuning achieve a similar maximum SM utilization at their maximum batch size.
% (3) The de-quantization kernel (dequant) is not sensitive to batch size, and achieves an equally high SM utilization with varying batch sizes.
% (4) Matrix multiplication (matmul) kernels achieve higher SM utilization with a larger batch size, as more data allows better utilization of the GPU's parallel processing resources.
% \textcolor{red}{Add a sentence here talking about what the main bottlenecks are and how can we further improve LLM fine-tuning performance moving forward (add more compute/add more memory bandwidth etc.?)...}

\textbf{Memory resource utilization study.}
Fig.~\ref{fig:mem_util} shows the kernel-level breakdown of GPU memory bandwidth utilization. 
We use the same experimental setup as in the evaluation of SM utilization, and find the following insights.
(1) For both sparse and dense fine-tuning, the time-weighted memory utilization decreases with increasing batch size. 
This is because the model parameters are loaded once and shared by all queries in a batch. 
However, a larger batch increases the execution time (as discussed in ~\S\ref{sec:sensitivity}), leading to a lower average memory bandwidth utilization. 
(2) For the same batch size, sparse fine-tuning achieves higher memory bandwidth utilization than dense fine-tuning due to shorter execution times.
(3) Dequant layers' memory utilization is batch-size-independent, while matmul layers' utilization decreases with larger batch sizes. To maximize GPU memory usage, a sufficiently large batch size should be used. With large batch sizes, fine-tuning becomes compute-bound, indicating a need for improved compute resources in future hardware to better utilize memory bandwidth.
% (2) With the same batch size, sparse fine-tuning achieves higher memory bandwidth utilization compared to dense fine-tuning, due to the shorter execution time.
% (3) At the kernel level, dequant layers' memory utilization is not sensitive to the batch size, matmul layers' memory utilization decreases as batch size increases.
% In practice, one should use a batch size that is large enough to use all available GPU memory. 
% With a large enough batch size, LLM fine-tuning is compute-bound, meaning there is not enough computing capability to fully utilize the memory bandwidth. 
% Future hardware development should aim to increase the compute resources to improve utilization.

\begin{tcolorbox}[width=0.48\textwidth]
\textbf{\OBFIVE.}
As the batch size increases, LLM fine-tuning transitions from being memory-bound to compute-bound.
\end{tcolorbox}

% SM utilization, Memory footprint, Memory utilization. Similar to runtime breakdown, this part is dataset-independent.

% \GREEN{A figure of SM util for Mixtral}

% \GREEN{A figure of SM util for BlackMamba}

% \GREEN{A figure of Mem util for Mixtral}

% \GREEN{A figure of Mem util for BlackMamba}

% \begin{itemize}
%     \item \BLUE{layer level breakdown: attention, MLP ...}
%     \item \BLUE{kernel level breakdown: dequant, matmul, silu ...}
%     \item \BLUE{compare high level, layer, kernel level SM and Mem util}
%     \item \BLUE{compare sparse with dense}
%     \item \BLUE{compare diff. batch size}
%     \item \BLUE{compare mixtral with mamba}
% \end{itemize}

% Explain results and add insights

\subsubsection{Effect of Load Imbalance Due to Fine-Tuning} \label{sec:load_moe}
\begin{figure}[t] 
\centering
\includegraphics[width=\linewidth,trim=1cm 6.5cm 1cm 0.5cm]{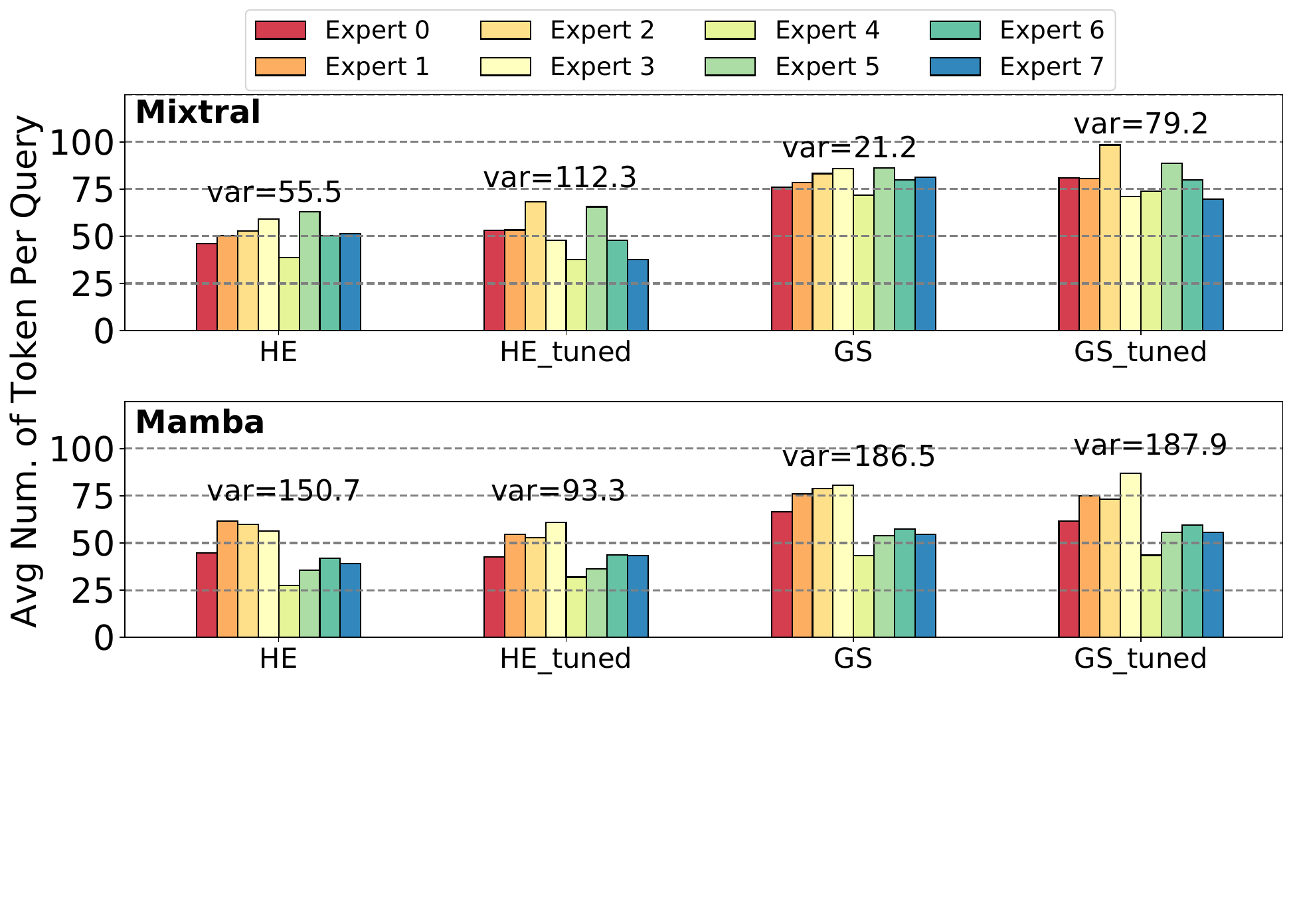}
\caption{Token distribution to different experts.}
\label{fig:token_dis}
\end{figure}

\begin{figure}[t] 
\centering
\includegraphics[width=\linewidth,trim=1cm 1cm 1cm 0.5cm]{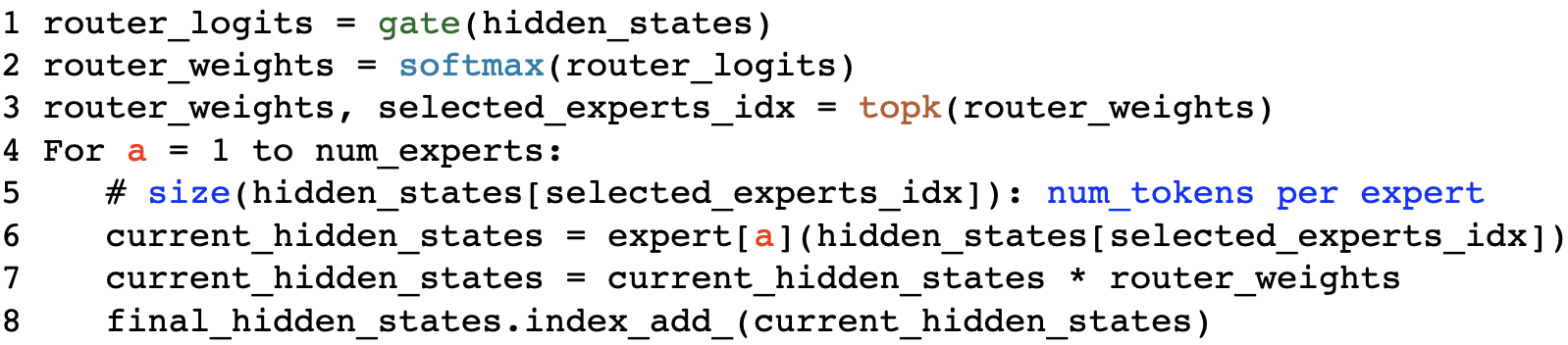}
\caption{Pseudo code for MoE layers.}
\label{fig:pseudo_moe}
\end{figure}
Recent trends in deploying expert parallelism in MoE models have highlighted load-imbalanced computation among experts as a significant issue impacting inference and training efficiency~\cite{tutel}.
During the training process of MoE models, each token is dynamically assigned to the top-k experts based on routing scores. 
This strategy often leads to most tokens being assigned to a small number of experts, resulting in load imbalance and slower training. 
Additionally, some experts receive insufficient training, which degrades overall model performance~\cite{shazeer2017outrageously}.
A naïve approach to address this imbalance is to use token dropping and padding to ensure that the number of tokens assigned to each expert is equal~\cite{lepikhin2020gshard}.
However, this method sacrifices model quality or leads to wasted computation. In this section, we analyze how fine-tuning influences the token distribution among experts. 
We compare the token distribution of \mixtral and \mamba before and after fine-tuning to understand the impact of this process.

We extract 1,000 examples from both the \cs and \mt datasets to test the original models without tuning and the models after 10 epochs of tuning on these datasets. Fig.~\ref{fig:pseudo_moe} provides the pseudo code for MoE layers with top-k gating. 
In this process, the hidden states are first sent to the router of the MoE layer, which generates router logits. 
These logits determine the priority of each expert for each token. Based on the router score for each token, tokens are grouped together and sent to their assigned experts. This top-k routing strategy can lead to load imbalance if the model has not been pre-trained for balance.

Fig.~\ref{fig:token_dis} evidently shows that fine-tuning causes load imbalance in \mixtral for both datasets.
Comparing variance before and after fine-tuning (e.g., HE vs. HE\_tuned), the token assignment variance increased from 55 to 112 for \cs and from 21 to 79 for \gs. Expert 3 became the most frequently used and important expert post fine-tuning.
Conversely, there is a decrease in the variance of token distribution for \mamba on the \cs dataset, dropping from 150 to 93.
For the \gs dataset, the token distribution variance for \mamba remains almost unchanged after fine-tuning.
This suggests that load-imbalance has a less disruptive impact on fine-tuning for \mamba compared to \mixtral.
From Fig.~\ref{fig:token_dis}, we can also observe that \mixtral demonstrates better load balance in both tasks compared to \mamba, despite the increased load imbalance after fine-tuning.
% \textcolor{red}{Add a sentence/two about how to address this challenge to further improve performance? Also, add a take-away message.}
% \GREEN{
The increased level of imbalance after fine-tuning suggests GPU load balancing techniques can be helpful. 
Both single GPU load balancing~\cite{chen2010} and multi-GPU load balancing~\cite{wahib2024elastic} have been proposed to address this issue.
% }

\begin{tcolorbox}[width=0.48\textwidth]
\textbf{\OBSIX.}
The effect of fine-tuning on expert load imbalance in the MoE layer is LLM model and dataset dependent.
\end{tcolorbox}

\subsubsection{Sensitivity Study on Sequence Length}
\label{sec:sensitivity}
% \begin{figure}[t]
% \centering
% \includegraphics[width=\linewidth,trim=1cm 4cm 1cm 0.5cm]{figures/ablation_latency.pdf}
% \caption{Sensitivity Study of Latency. }
% \label{fig:ablation_latency}
% \end{figure}
% \begin{figure}[t]
% \centering
% \includegraphics[width=\linewidth,trim=1cm 4.5cm 1cm 0.5cm]{figures/ablation_sm.pdf}
% \caption{Sensitivity Study of GPU Streaming Multi-processor (SM) utilization. }
% \label{fig:ablation_sm_util}
% \end{figure}
% \begin{figure}[t]
% \centering
% \includegraphics[width=\linewidth,trim=1cm 5cm 1cm 0.5cm]{figures/ablation_mem.pdf}
% \caption{Sensitivity Study of GPU DRAM utilization. }
% \label{fig:ablation_mem_util}
% \end{figure}

To further analyze the effect of sequence length on the fine-tuning process, we chose the batch size that would maximize the memory for each sequence length (64, 128, 256, 512, and 1024) and compared the latency, SM utilization, and DRAM utilization. 
Our evaluation (the figure is omitted from the paper due to page limitation) shows that the latency for \mixtral remains almost constant across different sequence lengths, while \mamba fine-tuning exhibited a slight reduction in latency as sequence length increased, with approximately 19\% and 25\% decreases for sparse and dense fine-tuning, respectively.
This is due to the varying maximum batch sizes supported by each sequence length, resulting in a similar number of tokens in each batch. 
Because latency remains consistent with increasing sequence length and we can use larger batch sizes, throughput is higher for shorter sequences.
\section{Analytical Model to Estimate the Cost of Fine-Tuning LLMs} \label{sec:analytical}

While training LLMs from scratch is a cost-prohibitive process, fine-tuning LLMs offers an attractive solution to align LLMs to desired behaviors.
One such example is fine-tuning LLMs to a domain-specific use-cases, for example, to answer math questions.
\S\ref{sec:accuracy} shows that it is possible to fine-tune pre-trained LLMs on domain-specific tasks to significant improve accuracy.
While this is a desired approach, currently, no model exists that can predict the cost of fine-tuning LLMs.

Fine-tuning LLMs is complex, influenced by factors like model size, GPU memory, dataset sequence length, and MoE sparsity, all affecting batch size and throughput. By integrating these factors with GPU costs, we can identify the most cost-efficient GPU for pre-tuning tasks. This section presents an analytical model based on previous characterization.
% Fine-tuning LLM is a convoluted task. 
% There are many parameters in the fine-tuning setup that can affect the cost efficiency. 
% One needs to consider the model size, GPU memory size, dataset sequence length, and the MoE sparsity, these parameters will affect the maximum batch size in fine-tuning, and the batch size will affect the throughput. 
% Combined with the GPU cost, we can find the most cost-efficient GPU for certain pre-tuning tasks. 
% In this section, we build an analytical model and evaluate the model based on the characterization performed in the former section.

This model estimates cloud-based fine-tuning costs for a given dataset and LLM. Developed from previous sections, it can be adapted for other LLMs by adjusting parameters. It assumes using the maximum batch size supported by GPU memory to optimize cost. We first estimate this batch size, then use it to evaluate throughput and fine-tuning costs.
% The goal of this model is to estimate the cost of cloud-based fine-tuning given an input dataset and an LLM model.
% Following the study in previous sections, we develop and correlate an analytical model for the LLMs evaluated in this paper.
% However, the proposed model is broadly applicable to other LLMs by changing model-related parameters.
% This model assumes that the user uses a maximum batch size supported by a given GPU memory capacity to optimize cost.
% Therefore, we first build a model to estimate the maximum batch size, and then incorporate this number into evaluating the throughput and cost of fine-tuning LLMs.

\subsection{Estimating Maximum Batch Size}
The maximum batch size is the maximum number of queries that can fit in GPU memory at once. 
Our analytical model for maximum batch size is shown in \eqref{eq:max_bs}.
\begin{equation}
\label{eq:max_bs}
    Max\_BSZ = \lfloor C_0*\frac{GPU\_mem - model\_mem}{seq\_len * ((1 - C_1) + C_1 * sparsity)} \rfloor
\end{equation}

Intuitively, larger GPU memory allows for higher batch sizes. 
In the meantime, the LLM model will take up a certain amount of GPU memory, and need to be subtracted in the analytical model. Fig.~\ref{fig:throughput} supports this by showing that on the same dataset, \mamba can support larger batch size than \mixtral because of its smaller model size.

Moreover, the sequence length and sparsity also affect the maximum batch size. 
Because the sparsity only affects the MoE part of the LLM, we multiply its influence by $C_1$, which we call \textit{MoE coefficient}.
We apply the sequence length and the sparsity in the denominator as they are inversely related to batch size. 
Then, we multiply the result by $C_0$, the \textit{scaling coefficient}, which scales the batch size by a constant. The \textit{scaling coefficient} is different across LLM models, because different models have different architecture (\S\ref{sec:setup}), and generate different amounts of intermediate data for each query. 
The \textit{scaling coefficient} for \mamba is higher than that of \mixtral because it is a smaller model. 
Finally, we use floor to round it to the maximum integer.
% \textcolor{red}{Yuhan: add how our profiling led to these decisions for the batch size.}

% \RED{try C1=0.97 for mixtral, and C1=0.9 for mamba.}

The \textit{MoE coefficient} and \textit{scaling coefficient} vary across models.
These coefficients are independent of GPU microarchitectural parameters.
We find the maximum batch size for both LLM models on NVIDIA A40 (48GB), A100 (40GB), A100 (80GB), and H100 (80GB), and apply our model to find the optimal coefficients. 
For \mixtral, $C_0=82$ and $C_1=0.95$, and for \mamba, $C_0=83$ and $C_1=0.88$.
While we showcase these parameters for the models evaluated, \S\ref{subsection:model_generalization} discusses how to generalize this approach for other models.
% The user of this model is not expected to determine the values of these constants; given a finite number of available pre-trained models, this responsibility lies with the model designer.
% \textcolor{red}{Add a figure showing batch size trend, and re-write the above paragraph.}
% \textcolor{red}{Add a related paper, show how it was inaccurate, and how we correct this?}
% \textcolor{red}{Add a discussion on how to calculate coeffieicents/whose responsibility it is.}

\begin{figure}[t]
\centering
    \includegraphics[width=\linewidth]{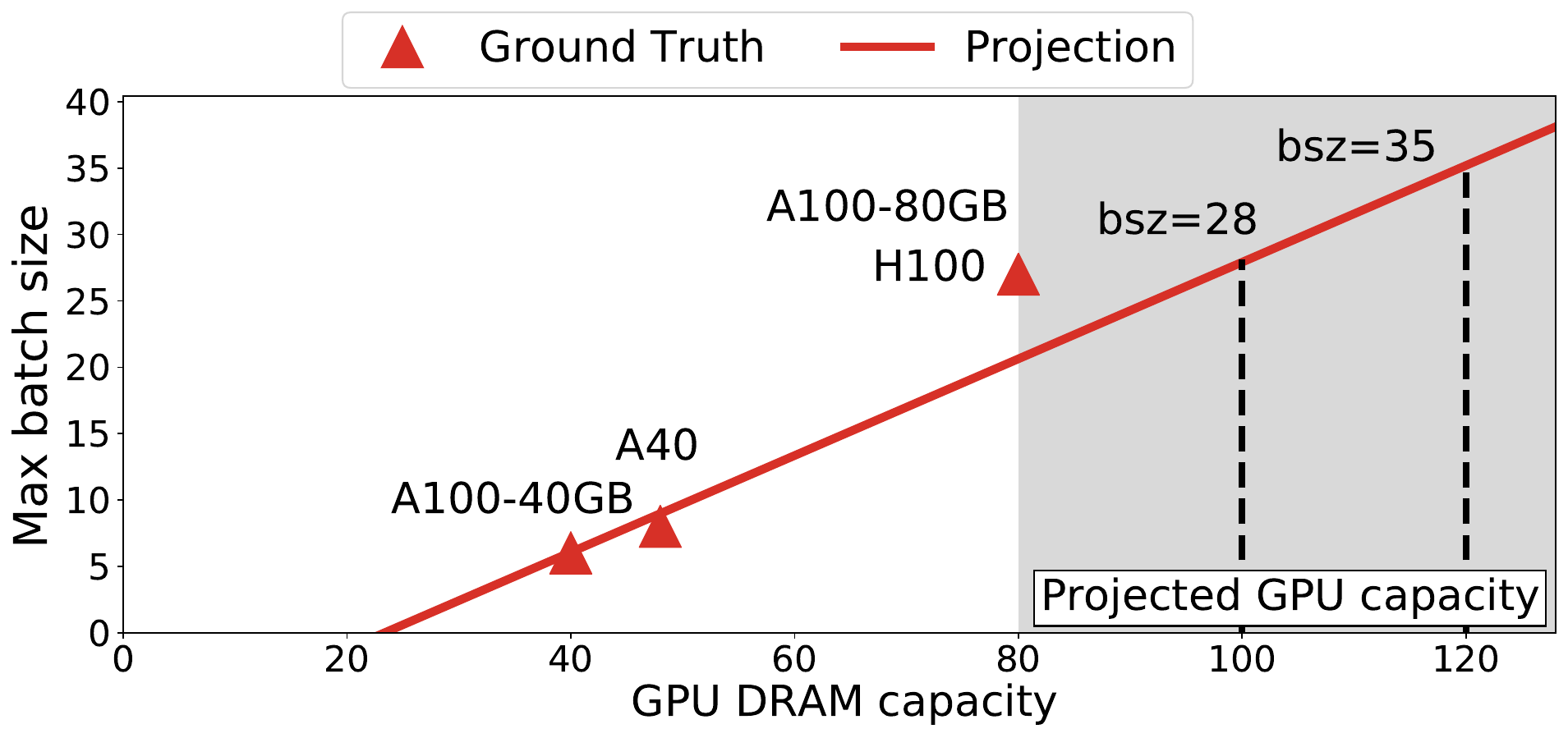}
\caption{Projected maximum batch size of Mixtral for different GPUs.}
\label{fig:analytical_capacity}
\end{figure}
Using our analytical model, we demonstrate the maximum batch sizes for fine-tuning on four different NVIDIA GPUs: A40, A100-40GB, A100-80GB and H100 with memory capacities of 48GB, 40GB, 80GB, and 80GB, respectively.
% We further experiment the maximum batch size for fine-tuning on three GPUs, NVIDIA A40, A100, and H100, which have 48GB, 40GB, and 80GB main memory respectively. 
Fig.~\ref{fig:analytical_capacity} shows our projected maximum batch size and correlate it with experimented ground truth.
While the maximum memory capacity available in NVIDIA GPUs today is 80GB, we use our analytical model to project the maximum batch size that future GPUs might support.
For GPU memory capacities of 100GB and 120GB, our model predicts that the maximum batch sizes supported for fine-tuning \mixtral will be 28 and 35, respectively.
Due to space limitations, we only show the projection of Mixtral model. 

\subsection{Estimating Throughput}
As discussed in \S\ref{sec:GPUcharacterization}, when the batch size increases, the LLM fine-tuning gradually switches from memory bound to compute bound. 
When the compute resources are abundant, the throughput increases almost linearly with batch size.
However, when compute resources become constrained, the throughput improvement gradually saturates. 
We model this behavior using a logarithmic relation between batch size and throughput.
% Therefore, the throughput grows logarithmically with respect to batch size.
Our analytical model for maximum batch size is shown in \eqref{eq:throughput_vs_bs}.
\begin{equation}
    \label{eq:throughput_vs_bs}
    Throughput = C_2 * log(\frac{batch\_size}{sparsity*C3}) + C_4
\end{equation}
In the equation, in addition to the basic logarithmic part, we have three coefficients $C_2$, $C_3$, and $C_4$. 
$C_2$ is the \textit{scaling coefficient}, which depends on the LLM model, GPU architecture, and the dataset. 
The higher the compute capability a GPU can provide, and the lower the LLM model and dataset compute requirement is, the higher the \textit{scaling coefficient} will be.
$C_3$ is the \textit{MoE attenuation coefficient}, which tunes how much the MoE sparsity affects the throughput. 
MoE sparsity only affects the MoE part in LLM model, and thus should be attenuated to avoid over compensation. 
This coefficient is only LLM model dependent, because once the model is fixed, the influence of sparsity is determined.
C4 is the \textit{intercept}, conceptually it equals to the throughput when batch size equals one, because the logarithmic part in \eqref{eq:throughput_vs_bs} is zero when batch size is one.
Using scipy~\cite{2020SciPy-NMeth} to fit the model and generate four sets $(C_2, C_3, C_4)$, for each model and dataset combination.

\begin{figure}[t]
\centering
    \includegraphics[width=\linewidth]{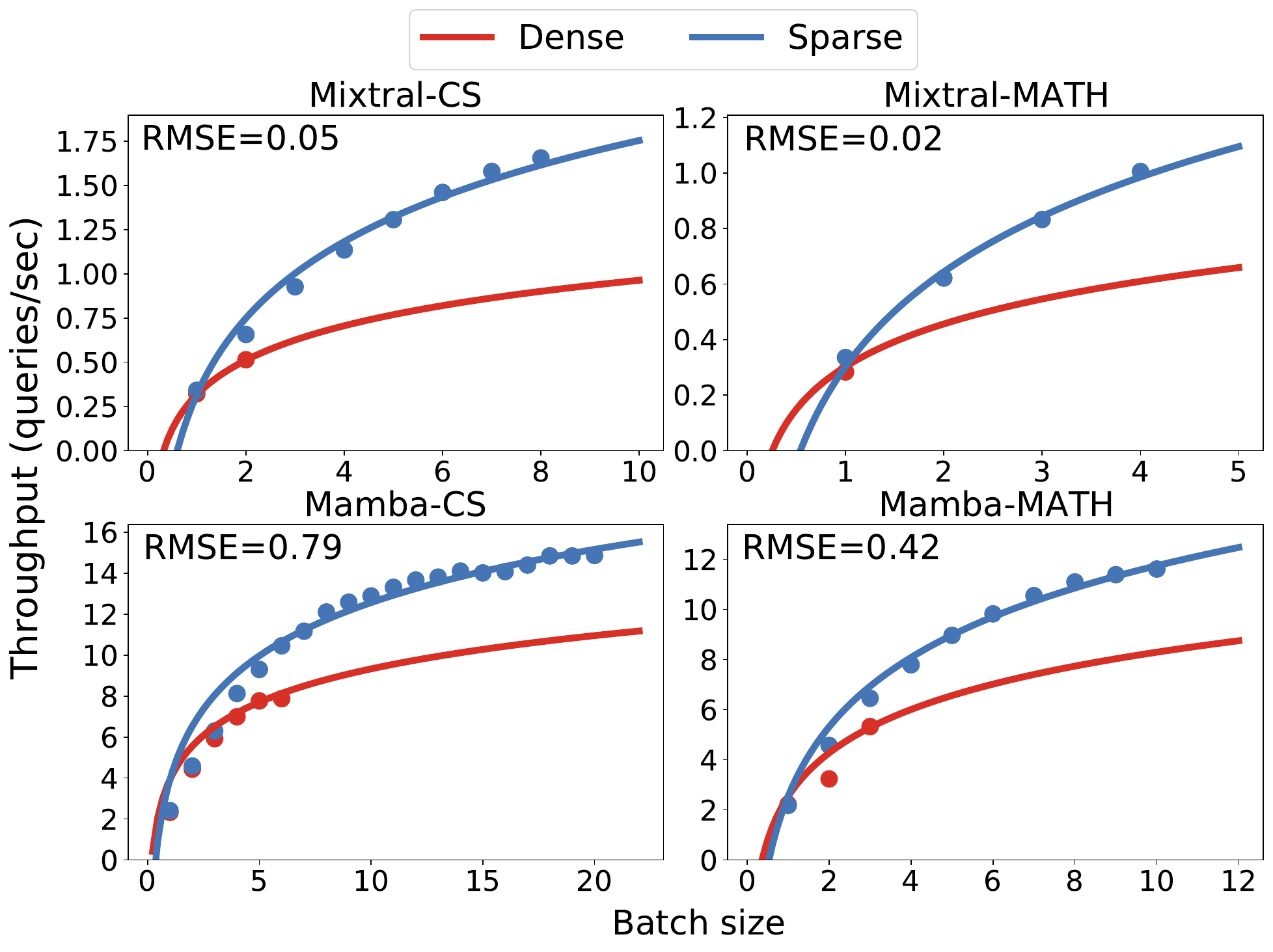}
\caption{Estimation and validation of LLM fine-tuning throughput for different models, datasets for A40 GPU. Dots represent ground truth and lines present the estimation.}
\label{fig:throughput_analytical}
\end{figure}

\begin{figure}[t]
\centering
    \includegraphics[width=\linewidth]{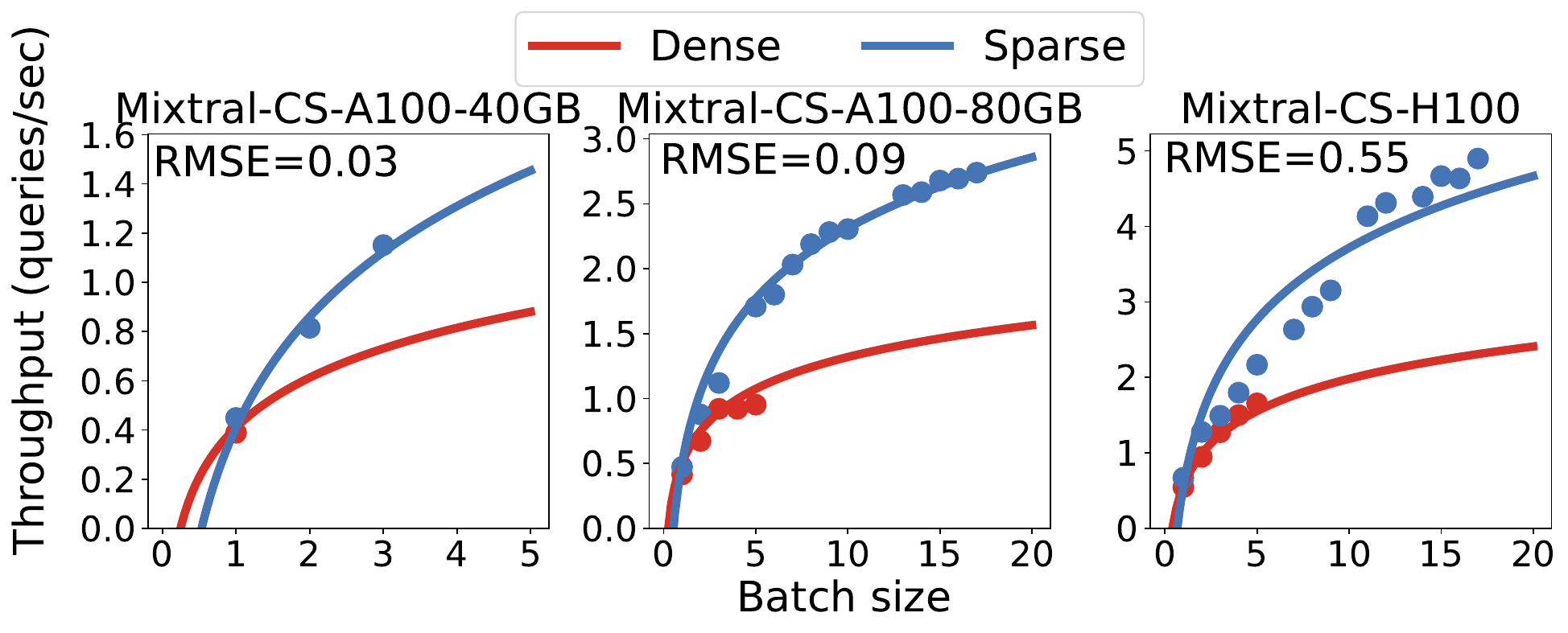}
\caption{Estimation and validation of fine-tuning throughput for \mixtral \gs for different GPUs: A100 and H100.}
\label{fig:analytical_gpu}
\end{figure}

To estimate the accuracy of this model, we correlate the model output with experimental data from our study.
Fig.~\ref{fig:throughput_analytical} shows this correlation study, where discrete data points (dots) represent experimental values, and the line represents output of our analytical model.
We use both dense and sparse \mixtral and \mamba for both datasets used in our study.
The figure clearly shows that our model accurately predicts LLM fine-tuning throughput with a Root Mean Squared Error (RMSE) of less than 0.8.
Fig.~\ref{fig:analytical_gpu} shows the correlation study of the analytical model of three other GPUs, A100 (40GB), A100 (80GB), and H100. The RMSE is less than 0.6, close to that of A40.

\subsection{Estimating the Total Fine-Tuning Cost}
\begin{table}[t]
    \centering
    \caption{Estimated cost of fine-tuning \mixtral on \gs with sparse MoE based on our analytical model}
    \begin{tabular}{c|c|c|c|c|c}
         GPU & Mem & MBS & Throughput & Cost (\$/hr) & Cost (\$)  \\
         \hline \hline
         A40 & 48GB & 4 & 1.01 & 0.79 & 32.7 \\
         A100 & 80GB & 17 & 2.74 & 1.67 & 25.4 \\
         H100 & 80GB & 17 & 4.90 & 2.1 & 17.9 \\
    \end{tabular}
    \vspace{5pt}
    
    \label{tab:gpu_cost}
\end{table}

Using the throughput estimation, we calculate the cost of fine-tuning LLMs for different GPUs. 
The cost of GPU resource renting per hour is calculated based on CUDO compute~\cite{cudo}, as other popular cloud providers do not offer cost/hour rates for the NVIDIA A40 GPU. 
However, one can easily adjust the GPU renting cost per hour to estimate the cost on other clouds such as Amazon AWS~\cite{aws} or Lambda~\cite{lambda}.
Table~\ref{tab:gpu_cost} estimates the cost for fine-tuning \mixtral on the \mt dataset with a sparse setup, using 10 epochs on different GPUs for a realistic cost estimate.
Enterprises may use larger datasets for fine-tuning, such as, OpenOrca~\cite{mukherjee2023orca} and LaMini-instruction~\cite{wu2024laminilm} containing more than 2M queries. 
For OpenOrca, by scaling the cost by number of queries, our model predicts that the most cost-effective option to rent GPU resources on CUDO compute is NVIDIA H100 with a net cost of \$3460. 

\subsection{Generalization of the Analytical Model} \label{subsection:model_generalization}
The analytical models for estimating maximum batch size and throughput can be generalized to various LLM models and datasets. These models consider the characteristics of the LLM, dataset, and GPU. Specifically, the maximum batch size model combines GPU memory and LLM model size to determine available memory for input data, while dataset sequence length and LLM sparsity determine space needed per batch. In throughput estimation, based on the observation we made (\S\ref{sec:GPUcharacterization} Takeaway 5), GPU shifts from memory-bound to compute-bound as batch size increases. This characteristic generally applies to all GPUs due to the resource constraint, so the logarithmic relation between batch size and throughput persists. The sparsity in \eqref{eq:throughput_vs_bs} is model dependent, the influence of GPU, LLM model, and dataset are embedded in the coefficients $C_2$, $C_3$, and $C_4$ in \eqref{eq:throughput_vs_bs}.

The coefficients in \eqref{eq:max_bs} and~\eqref{eq:throughput_vs_bs} are dependent on GPU, LLM model, and dataset; however, the underlying models are generalizable to unseen GPU, LLM model, and datasets. 
Although it takes some effort to sweep batch sizes and collect throughput data points to fit our models, the benefits greatly outweigh the cost. Once the models are fit, our model can help choose the most cost-efficient GPU for fine-tuning LLM models, greatly saving resources and money.

% To calculate the pre-tuning cost, we use the price of Amazon Web Service (AWS) GPU instances \RED{(Add ref)}. 
% Table~\ref{tab:gpu_cost} shows the GPU specification and cost. 

% \BLUE{\mixtral, \mt, sparse, 10 epochs}

% \RED{Add a GPU cost table}

% \RED{Add a scatter figure of throughput vs. GPU cost}
\section{Related Works}

Parameter-Efficient Fine-Tuning (PEFT) has been widely adopted to fine-tune LLM model for specialized tasks\cite{dettmers2023qlora, hu2021lora,houlsby2019parameter,he-etal-2022-sparseadapter, liu2024dora, zhao2024galore, jiang2024mora}. 
% Among them, MoE models activate only a subset of the entire model for each input, allowing them to scale up the model size without incurring additional computation cost. 
MoE additioally train specialized experts for different areas and the dynamic selection of experts makes it possible to scale the fine-tuning workload to different experts in parallel. 
\cite{pan2024dense, dou2024loramoe, zhou2022mixtureofexperts, dai2024deepseekmoe} show that MoE models can improve the ability to process knowledge for specific tasks, while maintaining the world knowledge in LLM.
Kim \textit{et al.}~\cite{kim2024llmem} construct an analytical model to estimate GPU memory consumption for distributed fine-tuning. The model also provides insights into optimizing memory usage through tensor, model, and pipeline parallelism.
% This paper studies the performance and estimate the cost of LLM fine-tuning using two popular MoE models Mixtral~\cite{jiang2024mixtral} and BlackMamba~\cite{anthony2024blackmamba}. 

\section{Conclusions}
\label{section:Conclusion}
Fine-tuning LLMs is an attractive technique for tailoring modern language models using domain-specific knowledge in a cost-effective manner.
This paper delved into understanding the performance of fine-tuning MoE LLM models on a single GPU.
Our profiling demonstrated that sparse MoE layers offer the best bang-for-buck trade-off.
Using our profiling results, we developed and validated an accurate analytical model to estimate the cost of LLM fine-tuning.
Using this model, we showed the dollar amount that needs to be budgeted for fine-tuning LLMs, which is much lower than pre-training.
For example, our model predicted that fine-tuning a sparse \mixtral model using a realistic data size of 2M queries can be done with NVIDIA H100 GPU with a cost of \$3460.
A way to further reduce cost based on our study is to add compute resources to accelerate the MoE layers.
While we showcase our study on fine-tuning LLMs using a single GPU, extending this model to multi-GPU systems is left for future exploration.

% \RED{TODO: check the following terms are consistent in the paper: Mixtral8x7, BlackMamba, fine-tuning/finetuning, GSM8K, Hellaswag, LoRA, QLoRA, MoE, LLaMA, Math\_14k, Commonsense\_15k.}

% \RED{TODO: double check the format requirement, no page number? comment out pagestyle\{plain\} to remove page number}

% \RED{TODO: training and fine-tuning are carelessly used in the paper, make them consistent.}

% \RED{Analytical model: memory, cost in \$, corelate with real-world numbers (find real LLM fine-tune cost example). 
% \\
% Input params: GPU mem size, model size, computation op count, dataset seq length, batchsize. 
% \\
% Output: exe time, cost}

\section*{Acknowledgments}
This work was supported in part by Semiconductor Research Corporation (SRC). We thank all the anonymous reviewers for their valuable comments and suggestions.

% The preferred spelling of the word ``acknowledgments'' in America is without 
% an ``e'' after the ``g''. Avoid the stilted expression ``one of us (R. B. 
% G.) thanks $\ldots$''. Instead, try ``R. B. G. thanks$\ldots$''. Put sponsor 
% acknowledgments in the unnumbered footnote on the first page.

\clearpage
\bibliographystyle{unsrt}
\bibliography{IEEEexample}
\clearpage
% LaTeX template for Artifact Evaluation V20240722
%
% Prepared by Grigori Fursin with contributions from Bruce Childers,
%   Michael Heroux, Michela Taufer and other colleagues.
%
% See examples of this Artifact Appendix in
%  * ASPLOS'24 "PyTorch 2: Faster Machine Learning Through Dynamic Python Bytecode Transformation and Graph Compilation": 
%      https://dl.acm.org/doi/10.1145/3620665.3640366
%  * SC'17 paper: https://dl.acm.org/citation.cfm?id=3126948
%  * CGO'17 paper: https://www.cl.cam.ac.uk/~sa614/papers/Software-Prefetching-CGO2017.pdf
%  * ACM ReQuEST-ASPLOS'18 paper: https://dl.acm.org/citation.cfm?doid=3229762.3229763
%
% (C)opyright 2014-2024 cTuning.org
%
% CC BY 4.0 license
%
% \documentclass{sigplanconf}

% \usepackage{hyperref}

% \begin{document}

%%%%%%%%%%%%%%%%%%%%%%%%%%%%%%%%%%%%%%%%%%%%%%%%%%%%
% When adding this appendix to your paper, 
% please remove above part
%%%%%%%%%%%%%%%%%%%%%%%%%%%%%%%%%%%%%%%%%%%%%%%%%%%%

\appendix
\section{Artifact Appendix}

%%%%%%%%%%%%%%%%%%%%%%%%%%%%%%%%%%%%%%%%%%%%%%%%%%%%%%%%%%%%%%%%%%%%%
\subsection{Abstract}

% {\em Obligatory}
{\em This artifact reproduces the results presented in the Characterization Study. It includes a detailed three-level runtime breakdown, analysis of SM and MEM utilization, and a comprehensive study of throughput.}

\subsection{Artifact check-list (meta-information)}

% {\em Obligatory. Use just a few informal keywords in all fields applicable to your artifacts
% and remove the rest. This information is needed to find appropriate reviewers and gradually 
% unify artifact meta information in Digital Libraries.}

{\small
\begin{itemize}
  % \item {\bf Algorithm: }
  % \item {\bf Program: }
  \item {\bf Compilation: }PyTorch
  % \item {\bf Transformations: }
  % \item {\bf Binary: }
  \item {\bf Model: }Mixtral-8x7B and BlackMamba-630M/2.8B
  \item {\bf Data set: }Hellaswag, GSM8k, MATH\_14k and commonsense\_15k (provided in GitHub reopsitory)
  \item {\bf Run-time environment: }Ubuntu 20.04.6
  \item {\bf Hardware: }NVIDIA A40 (48GB) GPU
  % \item {\bf Run-time state: }
  % \item {\bf Execution: }
  % \item {\bf Metrics: }
  \item {\bf Output: }Nsight Compute
  \item {\bf Experiments: }Fine-tune both models using different batch sizes and conduct a GPU characterization study
  \item {\bf How much disk space required (approximately)?: }100GB
  \item {\bf How much time is needed to prepare workflow (approximately)?: }1 hour
  \item {\bf How much time is needed to complete experiments (approximately)?: }Throughput and Runtime Breakdown experiments can be completed within 2 hours, while Nsight Compute profiling for SM and MEM utilization will take approximately 80 hours
  \item {\bf Publicly available?: }Yes
  % \item {\bf Code licenses (if publicly available)?: }
  % \item {\bf Data licenses (if publicly available)?: }
  \item {\bf Workflow framework used?: }LLaMA-Factory
  % \item {\bf Archived (provide DOI)?: }
\end{itemize}
}

%%%%%%%%%%%%%%%%%%%%%%%%%%%%%%%%%%%%%%%%%%%%%%%%%%%%%%%%%%%%%%%%%%%%%
\subsection{Description}

\subsubsection{How to access}
Our source code can be found at \href{https://github.com/stsxxx/finetune}{https://github.com/stsxxx/finetune}
% {\em Obligatory}

\subsubsection{Hardware dependencies}
\begin{itemize}

    \item We conducted all experiments on a server equipped with an Intel® Xeon® Platinum 8380 CPU @ 2.30GHz and an NVIDIA A40 (48GB) GPU
    \item Supported GPUs should have at least 48GB of memory and feature an Ampere architecture or newer
\end{itemize}

\subsubsection{Software dependencies}
\begin{itemize}
    \item A recent Linux release
    \item Python 3.8.10
    \item CUDA 11.8
    \item PyTorch 2.1.0 compatible with CUDA 11.8
    \item CUDA toolkit 11.8 
\end{itemize}

\subsubsection{Data sets}
Hellaswag, GSM8k, MATH\_14k and commonsense\_15k. We provide all of them in our GitHub repository.
\subsubsection{Models}
Mixtral-8x7B and BlackMamba-630M/2.8B. We provide the python script to download them from Huggingface. Mixtral-8x7B is a gated model, access request should be submitted here \href{https://huggingface.co/mistralai/Mixtral-8x7B-v0.1}{https://huggingface.co/mistralai/Mixtral-8x7B-v0.1}.
%%%%%%%%%%%%%%%%%%%%%%%%%%%%%%%%%%%%%%%%%%%%%%%%%%%%%%%%%%%%%%%%%%%%%
\subsection{Installation}

% {\em Obligatory}
For the Python environment, simply clone our repository and use conda to set up a new environment by running the following command:

\begin{Verbatim}[commandchars=\\\{\},formatcom=\color{black}]
\textcolor{darkgreen}{#create a new conda environment}
conda create --name=ft python=3.8
conda activate ft

\textcolor{darkgreen}{#install pytorch2.1.0+cu118}
conda install pytorch==2.1.0 \textbackslash 
torchvision==0.16.0 torchaudio==2.1.0 \textbackslash 
pytorch-cuda=11.8 -c pytorch -c nvidia

\textcolor{darkgreen}{#download the source code}
git clone https://github.com/stsxxx
/finetune.git
cd finetune

\textcolor{darkgreen}{#install all other dependencies}
pip install -r requirements.txt
\end{Verbatim}
%%%%%%%%%%%%%%%%%%%%%%%%%%%%%%%%%%%%%%%%%%%%%%%%%%%%%%%%%%%%%%%%%%%%%
\subsection{Experiment workflow}
First make sure the working directory is the LLaMA-Factory directory:

\begin{Verbatim}[commandchars=\\\{\},formatcom=\color{black}]
cd LLaMA-Factory 

\end{Verbatim}

Before running experiments, you should download both two models from Huggingface:

\begin{Verbatim}[commandchars=\\\{\},formatcom=\color{black}]
\textcolor{darkgreen}{#Add Blackmamba directory to your pythonpath}
export PYTHONPATH=$PYTHONPATH:../BlackMamba

\textcolor{darkgreen}{#specify where you want to store models}
export HF_HOME="path"

\textcolor{darkgreen}{#download models, huggingface access token}
\textcolor{darkgreen}{should be entered in the terminal}
python3 model_download.py

\end{Verbatim}

Make sure you change the transformers library path and model config file path before running each experiment bash script, you can find an example in the README file:
\begin{Verbatim}[commandchars=\\\{\},formatcom=\color{black}]
\textcolor{darkgreen}{# change it to your transformers library path} 
transformers_path="xxxxx"

\textcolor{darkgreen}{# change it to your model config path} 
config_file_path="xxxxx"

\end{Verbatim}

To reproduce the fine-tuning throughput results shown in Fig.~\ref{fig:throughput}, you can run the following scripts:
\begin{Verbatim}[commandchars=\\\{\},formatcom=\color{black}]
./mixtral_tp.sh
python3 throughput.py ./profile_data/mixtral
/throughput > mixtral_throughput.txt

./mamba_tp.sh
python3 throughput.py ./profile_data
/blackmamba/throughput > mamba_throughput.txt

\end{Verbatim}

High-level and layer-level latency breakdown results shown in Fig.~\ref{fig:latency_highlevel} and~\ref{fig:latency_layerlevel} can be obtained by running:
\begin{Verbatim}[commandchars=\\\{\},formatcom=\color{black}]
./mixtral_lt.sh

python3 mixtral_latency.py ./profile_data
/mixtral/latency > mixtral_latency_breakdown.txt

./mamba_lt.sh
python3 mamba_latency.py ./profile_data
/blackmamba/latency > mamba_latency_breakdown.txt

\end{Verbatim}

You can also use Nsight Compute to profile and generate kernel-level latency breakdown, SM and MEM utilization results shown in Fig.~\ref{fig:latency_kernellevel},~\ref{fig:sm_util} and~\ref{fig:mem_util} by running:
\begin{Verbatim}[commandchars=\\\{\},formatcom=\color{black}]
./mixtral_pf.sh
python3 sm_mixtral.py ./profile_data/mixtral
/ncu > mixtral_sm.txt
python3 mem_mixtral.py ./profile_data/mixtral
/ncu > mixtral_mem.txt

./mamba_pf.sh
python3 sm_mamba.py ./profile_data/blackmamba
/ncu > mamba_sm.txt
python3 mem_mamba.py ./profile_data/blackmamba
/ncu > mamba_mem.txt
python3 sm_mamba_back.py ./profile_data
/blackmamba/ncu_back > mamba_sm_backward.txt
python3 mem_mamba_back.py ./profile_data
/blackmamba/ncu_back > mamba_mem_backward.txt
\end{Verbatim}

%%%%%%%%%%%%%%%%%%%%%%%%%%%%%%%%%%%%%%%%%%%%%%%%%%%%%%%%%%%%%%%%%%%%%
\subsection{Evaluation and expected results}

% {\em Obligatory}
The generated results are stored in specific text files as indicated in the commands above, such as mixtral\_sm.txt for SM utilization data of the Mixtral model.
%%%%%%%%%%%%%%%%%%%%%%%%%%%%%%%%%%%%%%%%%%%%%%%%%%%%%%%%%%%%%%%%%%%%%
\subsection{Experiment customization}
Customized experiments can be conducted with varying batch sizes and query sequence lengths, both of which can be adjusted in each bash script.
%%%%%%%%%%%%%%%%%%%%%%%%%%%%%%%%%%%%%%%%%%%%%%%%%%%%%%%%%%%%%%%%%%%%%
% \subsection{Notes}

%%%%%%%%%%%%%%%%%%%%%%%%%%%%%%%%%%%%%%%%%%%%%%%%%%%%%%%%%%%%%%%%%%%%%
\subsection{Methodology}

Submission, reviewing and badging methodology:

\begin{itemize}
  \item \url{https://www.acm.org/publications/policies/artifact-review-and-badging-current}
  \item \url{https://cTuning.org/ae}
\end{itemize}

%%%%%%%%%%%%%%%%%%%%%%%%%%%%%%%%%%%%%%%%%%%%%%%%%%%%
% When adding this appendix to your paper, 
% please remove below part
%%%%%%%%%%%%%%%%%%%%%%%%%%%%%%%%%%%%%%%%%%%%%%%%%%%%

% \end{document}

\end{document}